\documentclass[10pt,twocolumn,letterpaper]{article}

\usepackage{cvpr}
\usepackage{times}

\usepackage{xcolor}
\usepackage{url}            % simple URL typesetting
\usepackage{amsfonts}       % blackboard math symbols
\usepackage{nicefrac}       % compact symbols for 1/2, etc.
\usepackage{microtype}      % microtypography
\usepackage{subcaption}
\usepackage{multirow}
\usepackage{amsmath}
\usepackage{mathtools}
\usepackage{verbatim}
\usepackage{amssymb}
\usepackage{colortbl}
\usepackage{xspace}
\usepackage{tikz}
\usepackage{afterpage}
\usepackage{enumitem}
\usepackage{placeins}
\usepackage{listings}
\usepackage[square,numbers,sort&compress]{natbib}
\renewcommand{\citealp}[1]{\citet{#1}}
\let\oldciteauthor=\citeauthor
\renewcommand{\citeauthor}[1]{{\protect\NoHyper\oldciteauthor{#1}\protect\endNoHyper}\xspace}
\DeclareCaptionLabelSeparator{dot}{.~}
\usepackage{caption}
\usepackage[pagebackref=true,breaklinks=true,colorlinks,bookmarks=false]{hyperref}
\usepackage{algpseudocode}

\usepackage{amsthm, amsmath, amssymb}
 
\theoremstyle{definition}

\usepackage{amsmath, latexsym}
\usepackage{amsfonts}
\usepackage{mathtools}
\usepackage{dsfont}
\usepackage[]{xcolor}
\usepackage{colortbl}
\usepackage[colorinlistoftodos]{todonotes}
\usepackage{booktabs}
\usepackage{xfrac}
\usepackage{bbm}

\usepackage{algorithm}
\usepackage{algorithmicx}

\usepackage[most]{tcolorbox}
\usepackage{xparse}
\usepackage{lipsum}
\usepackage{changepage}
\usepackage{enumitem}

\usepackage{scalerel,stackengine}
\stackMath

\newcommand{\squishlist}{
   \begin{list}{$\bullet$}
    { \setlength{\itemsep}{0pt}      \setlength{\parsep}{3pt}
      \setlength{\topsep}{3pt}       \setlength{\partopsep}{0pt}
      \setlength{\leftmargin}{1.5em} \setlength{\labelwidth}{1em}
      \setlength{\labelsep}{0.5em} } }

\newcommand{\squishlisttwo}{
   \begin{list}{$\bullet$}
    { \setlength{\itemsep}{0pt}    \setlength{\parsep}{0pt}
      \setlength{\topsep}{0pt}     \setlength{\partopsep}{0pt}
      \setlength{\leftmargin}{2em} \setlength{\labelwidth}{1.5em}
      \setlength{\labelsep}{0.5em} } }

\newcommand{\squishend}{
    \end{list}  }

\newcommand{\myvecsym}[1]{\boldsymbol{#1}}

\newcommand{\grad}{\nabla}
\newcommand{\proj}{\ensuremath{\mathtt{proj}}\xspace}

\newcommand{\vzero}{\myvecsym{0}}
\newcommand{\vone}{\myvecsym{1}}

\DeclareMathOperator*{\E}{\mathbb{E}}

\newcommand{\expect}[2]{\mathds{E}_{{#1}} \left[ {#2} \right]}

\DeclareMathAlphabet{\mathpzc}{OT1}{pzc}{m}{n}

\newcommand{\R}{\mathbb{R}}

\DeclareMathOperator*{\argmax}{argmax}
\DeclareMathOperator*{\argmin}{argmin}

\newcommand{\mixup}{\textit{mixup}\xspace}
\newcommand{\cutout}{\textit{Cutout}\xspace}
\newcommand{\cutmix}{\textit{CutMix}\xspace}
\newcommand{\autoaugment}{\textit{AutoAugment}\xspace}
\newcommand{\linf}{\ensuremath{\ell_\infty}\xspace}

\newcommand{\lp}{\ensuremath{\ell_p}\xspace}

\DeclareMathOperator*{\maximize}{max}

\newcommand{\mnist}{\textsc{Mnist}\xspace}
\newcommand{\coloredmnist}{Color-\textsc{Mnist}\xspace}

\newcommand{\svhn}{\textsc{Svhn}\xspace}
\newcommand{\imagenet}{\textsc{ImageNet}\xspace}
\newcommand{\celeba}{\textsc{CelebA}\xspace}
\newcommand{\stylegan}{\textit{StyleGAN}\xspace}
\newcommand{\decode}{\ensuremath{\mathtt{dec}}\xspace}
\newcommand{\encode}{\ensuremath{\mathtt{enc}}\xspace}
\newcommand{\map}{\ensuremath{\mathtt{map}}\xspace}
\newcommand{\vgg}{\ensuremath{\mathtt{vgg}}\xspace}

\newcommand{\methodname}{\emph{Adversarial Mixing with Disentangled Representations}\xspace}
\newcommand{\methodnamenormal}{Adversarial Mixing with Disentangled Representations\xspace}
\newcommand{\methodabbrv}{\emph{AdvMix}\xspace}
\newcommand{\randomabbrv}{\emph{RandMix}\xspace}

\definecolor{Highlight}{gray}{0.9}
\newcommand{\Tstrut}{\rule{0pt}{2.6ex}}
\newcommand{\Bstrut}{\rule[-0.9ex]{0pt}{0pt}}
\newcommand{\TBstrut}{\Tstrut\Bstrut}

\newcommand\reallywidehat[1]{%
\savestack{\tmpbox}{\stretchto{%
  \scaleto{%
    \scalerel*[\widthof{\ensuremath{#1}}]{\kern-.6pt\bigwedge\kern-.6pt}%
    {\rule[-\textheight/2]{1ex}{\textheight}}%WIDTH-LIMITED BIG WEDGE
  }{\textheight}% 
}{0.5ex}}%
\stackon[1pt]{#1}{\tmpbox}%
}

\definecolor{dmblue50}{HTML}{D9E2FF}
\definecolor{dmblue100}{HTML}{AABFFF}
\definecolor{dmblue300}{HTML}{2267EB}
\definecolor{dmblue400}{HTML}{0053D6}
\definecolor{dmblue500}{HTML}{0044CC}
\definecolor{dmblue600}{HTML}{123693}
\definecolor{dmblue700}{HTML}{14234B}
 
\definecolor{dmgray50}{HTML}{F0F5F5}
\definecolor{dmgray100}{HTML}{D8DDDD}
\definecolor{dmgray300}{HTML}{C0C4C4}
\definecolor{dmgray400}{HTML}{A8ACAC}
\definecolor{dmgray500}{HTML}{909393}
\definecolor{dmgray600}{HTML}{737676}
\definecolor{dmgray700}{HTML}{565858}

\definecolor{dmteal50}{HTML}{D4FFED}
\definecolor{dmteal100}{HTML}{A6FFE0}
\definecolor{dmteal300}{HTML}{49EBD0}
\definecolor{dmteal400}{HTML}{33E0C8}
\definecolor{dmteal500}{HTML}{14C8B9}

\definecolor{dmpurple50}{HTML}{F0EBFF}
\definecolor{dmpurple100}{HTML}{E6D6FF}
\definecolor{dmpurple300}{HTML}{AD70FF}
\definecolor{dmpurple400}{HTML}{A261FF}
\definecolor{dmpurple500}{HTML}{6932E6}

\definecolor{dmred50}{HTML}{FFE6F0}
\definecolor{dmred100}{HTML}{FFD4E4}
\definecolor{dmred300}{HTML}{FF617B}
\definecolor{dmred400}{HTML}{F9526B}
\definecolor{dmred500}{HTML}{E1144B}

\definecolor{dmorange50}{HTML}{FFEACC}
\definecolor{dmorange100}{HTML}{FFDEAF}
\definecolor{dmorange300}{HTML}{FF8A4F}
\definecolor{dmorange400}{HTML}{FF7D45}
\definecolor{dmorange500}{HTML}{FF5F19}

\definecolor{dmyellow300}{HTML}{FFF26E}
\definecolor{dmyellow500}{HTML}{FFDB13}

\definecolor{dmlime50}{HTML}{F6FED8}
\definecolor{dmlime100}{HTML}{EDFCB3}
\definecolor{dmlime300}{HTML}{D6F36D}
\definecolor{dmlime400}{HTML}{CAEC4D}
\definecolor{dmlime500}{HTML}{B2D141}

\definecolor{dmpink50}{HTML}{F8D9FB}
\definecolor{dmpink100}{HTML}{F5C7F8}
\definecolor{dmpink300}{HTML}{E380EA}
\definecolor{dmpink400}{HTML}{D85FE1}
\definecolor{dmpink500}{HTML}{AE50BB}

\definecolor{dmcyan50}{HTML}{DAF6FF}
\definecolor{dmcyan100}{HTML}{C8F1FF}
\definecolor{dmcyan300}{HTML}{84DCFF}
\definecolor{dmcyan400}{HTML}{55CAF9}
\definecolor{dmcyan500}{HTML}{48B2DC}

\usepackage{tikz}
\usetikzlibrary{shapes,arrows,shadows,fit, decorations.pathmorphing}
\tikzstyle{mathop} = [draw, fill=red!20, text centered, minimum height=2em, circle]
\tikzstyle{op} = [draw, text width=6em, fill=blue!20, text centered, minimum height=2em, rounded corners]
\tikzstyle{none} = [text centered, minimum height=2em]
\tikzstyle{system} = [draw, dotted, minimum height=2em]
\tikzstyle{null} = [inner sep=0, outer sep=0]
\tikzset{>=latex}
\tikzset{every picture/.style={line width=1pt}}

\graphicspath{{compressed_figures/}}

\definecolor{codegreen}{rgb}{0,0.6,0}
\definecolor{codegray}{rgb}{0.5,0.5,0.5}
\definecolor{codepurple}{rgb}{0.58,0,0.82}
\definecolor{backcolor}{rgb}{0.95,0.95,0.92}
\lstdefinestyle{mystyle}{
    backgroundcolor=\color{backcolor},
    commentstyle=\color{codegreen},
    keywordstyle=\color{magenta},
    numberstyle=\tiny\color{codegray},
    stringstyle=\color{codepurple},
    basicstyle=\ttfamily\footnotesize,
    breakatwhitespace=false,
    breaklines=true,
    captionpos=b,
    keepspaces=true,
    numbers=left,
    numbersep=5pt,
    showspaces=false,
    showstringspaces=false,
    showtabs=false,
    tabsize=2
}
\makeatletter
\def\bstctlcite{\@ifnextchar[{\@bstctlcite}{\@bstctlcite[@auxout]}}
\def\@bstctlcite[#1]#2{\@bsphack
  \@for\@citeb:=#2\do{%
    \edef\@citeb{\expandafter\@firstofone\@citeb}%
    \if@filesw\immediate\write\csname #1\endcsname{\string\citation{\@citeb}}\fi}%
  \@esphack}
\makeatother

\cvprfinalcopy % *** Uncomment this line for the final submission

\def\cvprPaperID{9412} % *** Enter the CVPR Paper ID here

\ifcvprfinal\fi

\begin{document}

%\title{Adversarial Mixing with Disentangled Representations \\ for Invariant Risk Minimization}

\title{Achieving Robustness in the Wild via \\
Adversarial Mixing with Disentangled Representations}

% Improving generalization in image classification via adversarial mixing with disentagled representations
% Improving generalization in image classification via semantic adversarial training
% Express these variations through example.
% Data-driven specifications formulation => technique specific to STyleGAN

% Data-driven specifications for expressing and enforcing robustness to real-world variations
% Adversarial Training over Disentangled Latents
% Invariance to Data Driven Specification
%Expressing and enforcing robustness to input variations.
% Adversarially Invariant Risk Minimization

% Disentangled - Data-driven - Example - Adversarial - Robustness - variations, augmentation

% Beyond analytical specification
% Adversarial Mixing with Disentangled Representations [for invariance risk minimization]
% AdvMix

\author{Sven Gowal$^*$ \hspace{-1.2cm} \\
DeepMind  \hspace{-1.2cm} \\
{\tt\small sgowal@google.com} \hspace{-1.2cm}
\and
Chongli Qin$^*$\\
% DeepMind\\
{\tt\small chongliqin@google.com} \hspace{-.5cm}
\and
Po-Sen Huang\\
% DeepMind\\
{\tt\small posenhuang@google.com} \hspace{-.5cm}
\and
Taylan Cemgil\\
% DeepMind\\
{\tt\small taylancemgil@google.com}
\and
Krishnamurthy (Dj) Dvijotham\\
% DeepMind\\
{\tt\small dvij@google.com}
\and
Timothy Mann\\
% DeepMind\\
{\tt\small timothymann@google.com}
\and
Pushmeet Kohli\\
% DeepMind\\
{\tt\small pushmeet@google.com}
}
\maketitle

\begin{abstract}
Recent research has made the surprising finding that state-of-the-art deep learning models sometimes fail to generalize to small variations of the input.
Adversarial training has been shown to be an effective approach to overcome this problem.
However, its application has been limited to enforcing invariance to analytically defined transformations like \lp-norm bounded perturbations.
Such perturbations do not necessarily cover plausible real-world variations that preserve the semantics of the input (such as a change in lighting conditions).
In this paper, we propose a novel approach to express and formalize robustness to these kinds of real-world transformations of the input.
The two key ideas underlying our formulation are (1) leveraging disentangled representations of the input to define different factors of variations, and (2) generating new input images by adversarially composing the representations of different images.
We use a \stylegan model to demonstrate the efficacy of this framework.
Specifically, we leverage the disentangled latent representations computed by a \stylegan model to generate perturbations of an image that are similar to real-world variations (like adding make-up, or changing the skin-tone of a person) and train models to be invariant to these perturbations.
Extensive experiments show that our method improves generalization and reduces the effect of spurious correlations (reducing the error rate of a ``smile'' detector by 21\% for example).
\end{abstract}

\section{Introduction}

The principle by which neural networks are trained to minimize their average error on the training data is known as Empirical Risk Minimization (ERM) \citep{vapnik1998statistical}.
ERM has, for the most part, enabled breakthroughs in a wide variety of fields~\citep{goodfellow_deep_2016,krizhevsky_imagenet_2012,hinton_deep_2012}, and this success has lead to the usage of neural networks in applications that are safety-critical \citep{julian_policy_2016}.
ERM, however, is only guaranteed to produce meaningful models when the data encountered during training and deployment is drawn independently from the same distribution.
When a mismatch between training and testing data occurs, models can fail in catastrophic ways; and, unfortunately, such occurrence is commonplace: training data is often collected through a biased process that highlights confounding factors and spurious correlations~\cite{torralba2011unbiased, kuehlkamp2017gender}, which can lead to undesirable consequences (e.g., {\small\url{http://gendershades.org}}).

\begin{figure}[t]
\centering
\includegraphics[width=\columnwidth]{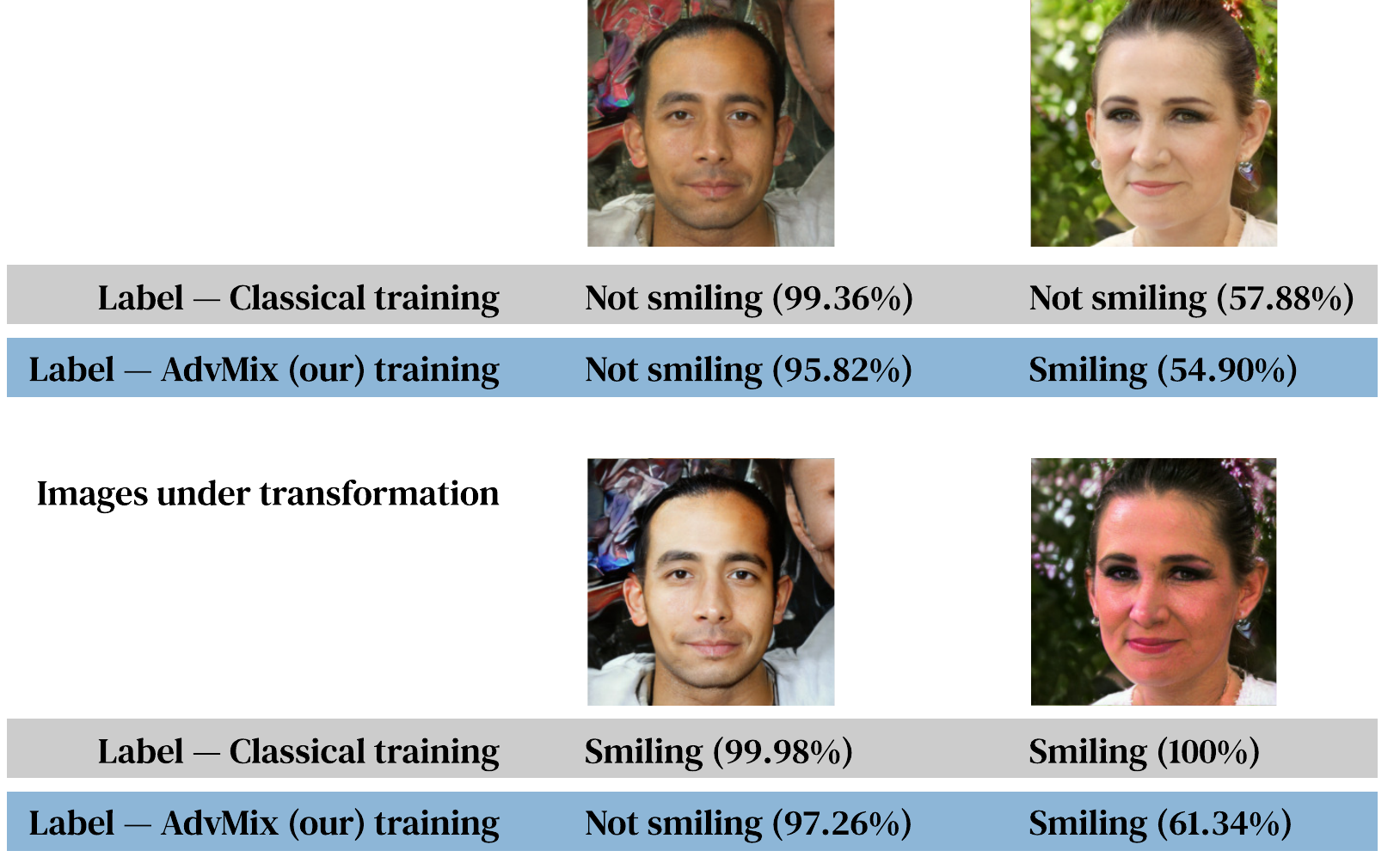}
\caption{
Variations of the same faces. A model obtained through classical training classifies the same face as both ``smiling'' and ``not smiling'' (depending on the variations). Our model remains consistent in terms of classification. Note that these persons ``do not exist'' and have been generated using a \stylegan model.
}
\label{fig:example}
\end{figure}

The effects of such data shifts are largely detailed in the literature.
For example, both \citet{recht2019imagenet} and~\citet{hendrycks2019natural} show that the accuracy of \imagenet models is severely impacted by changes in the data collection process.
Methods to counteract such effect, which mainly consist of \emph{data augmentation} techniques, also struggle.
Training against corrupted data only forces the memorization of such corruptions and, as a result, these models fail to generalize to new corruptions~\cite{vasiljevic2016examining,geirhos2018generalisation}.
Works such as \mixup~\cite{zhang2017mixup} or \autoaugment~\cite{cubuk2018autoaugment} pave the way to further improvements, but still require intricate fine-tuning to succeed in practice.

Another parallel and important line of work uncovered that the addition of small but carefully chosen deviations to the input, called adversarial perturbations, can cause the neural network to make incorrect predictions with high confidence \citep{carlini_adversarial_2017,carlini_towards_2017,goodfellow_explaining_2014,kurakin2016adversarial,szegedy_intriguing_2013}.
Techniques to build models that are robust to adversarially perturbed examples, such as adversarial training~\cite{madry_towards_2017}, have received a significant amount of attention in the recent years~\citep{goodfellow_explaining_2014,papernot_distillation_2015,kannan_adversarial_2018,xie_feature_2018}.
The existence of imperceptible perturbations that alter a model's output demonstrates that supervised learning algorithms still fail to capture the true causal relationships between signal and label.
The degradation of performance occurred when shifting between training and adversarial (or otherwise corrupted) distributions indicates that neural networks pick up on correlations that are not necessarily robust to small input perturbations~\citep{ilyas2019adversarial}.
The existence of imperceptible adversarial perturbations highlights just one form of spurious correlation that causes undesirable behaviors in the networks we train.

This paper focuses on training models that are robust to plausible real-world perturbations that preserve semantic content (such as those presented in Figure~\ref{fig:example}).
We go beyond conventional \emph{data augmentation} and \emph{adversarial training} on $l_p$-norm bounded perturbations by leveraging high-quality generative models that can describe such perturbations.
In particular, we address the question: ``Given a generative model with a sufficiently good disentangled representation that aligns well with the perturbations of interest, can we train neural networks that are resistant to bias and spurious correlations present in the training data?''
More specifically, we consider \stylegan~\citep{karras2019style} as our underlying generative model. Our contributions are as follows:
\begin{enumerate}[noitemsep]
    \item We develop a framework dubbed \methodname (\methodabbrv) which leverages the disentangled latents of a generative model to train networks that are robust to real-world variations.
    \item We demonstrate how to leverage \emph{StyleGAN}'s \emph{mixing} property to systematically transfer image attributes likely to be misclassified across image instances, thus allowing us to generate realistic worst-case semantic variations. This enables us to define semantic perturbations in a purely data-driven fashion, as opposed to methods that require data collection under different conditions \citep{arjovsky2019invariant}.
    \item We conduct extensive experiments on a controlled \coloredmnist dataset that compare \methodname with \emph{random data augmentation} and demonstrate under which conditions \methodabbrv achieves higher accuracy.
    \item Finally, we demonstrate empirically on \celeba that accuracy is not necessarily at odds with robustness~\citep{tsipras2018robustness}, once we consider semantic variations other than \lp-norm bounded variations.
\end{enumerate}

\section{Related work}

\paragraph{Robustness to \lp-norm perturbations.}

Generating pixel-level adversarial perturbations has been and remains extensively studied~\citep{goodfellow_explaining_2014,kurakin_adversarial_2016,szegedy_intriguing_2013,moosavi-dezfooli_robustness_2018,papernot_distillation_2015,madry_towards_2017}.
Most works focus the robustness of classifiers under \lp-norm bounded perturbations.
In particular, it is expected that a \emph{robust} classifier be invariant to small perturbations in the pixel space (as defined by the \lp-norm).
\citet{goodfellow_explaining_2014} and \citet{madry_towards_2017} laid down foundational principles to train robust networks, and recent works \citep{zhang2019theoretically, qin2019adversarial} continue to find novel approaches to enhance robustness.
While existing work is able to train models that are robust to imperceptible pixel-level variations, the study of robustness against semantically meaningful perturbations is largely under-explored.

\paragraph{Adversarial robustness beyond \lp-norm.}

\citet{engstrom2017rotation} and \citet{kanbak2018geometric} explored geometric transformations such as rotations and translation of images.
Early works (e.g., \citealp{baluja_adversarial_2017}) also demonstrated that it is possible to go beyond analytically defined variations by using generative models to create perturbations.
\citet{song2018constructing} and \citet{xiao_generating_2018} used a pre-trained AC-GAN~\citep{odena2017conditional} to generate perturbations; and they demonstrated that it is possible to generate semantically relevant perturbations for tasks such as \mnist, \svhn and \celeba.
Lastly, \citet{qiu2019semanticadv} have attempted to generate adversarial examples by interpolating through the attribute space defined by a generative model.
With the exception of \cite{jalal2017robust}, in which the authors strongly limit semantic variations by keeping the perturbed image close to its original counterpart, there has been little to no work demonstrating robustness to large semantically plausible variations.
As such the effect of training models robust to such variations is unclear.
To the best of our knowledge, this paper is the first to analyze the difference between \emph{adversarial training} and \emph{data augmentation} in the space of semantically meaningful variations.

\paragraph{Data augmentation}

\begin{figure}[t]
\centering
\includegraphics[height=2.5cm]{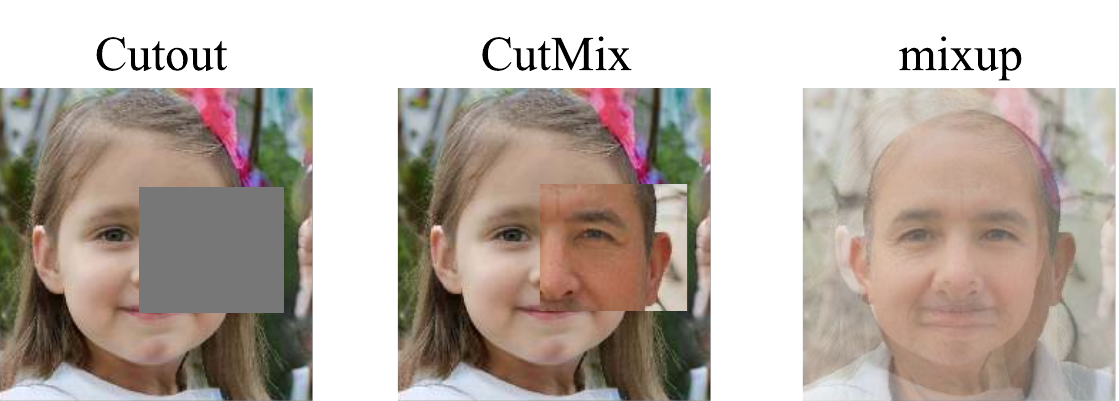}
\caption{
Comparison of different data augmentation techniques. These transformations tend to destroy the image semantics.
}
\label{fig:augmentation_example}
\end{figure}

Data augmentation can reduce generalization error.
For image classification tasks, random flips, rotations and crops are commonly used~\cite{he2016deep}.
More sophisticated techniques such as \emph{Cutout}~\cite{devries2017improved} (which produces random occlusions), \emph{CutMix}~\cite{yun2019cutmix} (which replaces parts of an image with another) and \emph{mixup}~\cite{zhang2017mixup} (which linearly interpolates between two images) all demonstrate extremely compelling and surprising results.
Indeed, while these methods often result in images that are visibly corrupted and void of semantic meaning (even to the human eye), the resulting models often achieve state-of-the-art accuracy across a wide range of datasets.
Figure~\ref{fig:augmentation_example} shows a comparison of these different techniques.
Some of these data augmentation techniques have been applied to latent representations of the input (rather than the input itself)~\cite{verma2018manifold}.
However, these do not focus on the effect of data bias.

\paragraph{Causal reasoning using additional data.}

\citet{heinze2017conditional} use grouped observations (e.g., the same object under different conditions) to discover variations that should not explain the classification label.
More recently \citet{arjovsky2019invariant} developed a method called Invariant Risk Minimization (IRM) which tries to find an invariant predictor across different environments (or groups of data points). 
Both methods were able to build classifiers that were less sensitive to spurious correlations, which, in turn, lead to classifiers that were less biased than classifiers trained purely on an original biased training set.
However, they require explicitly annotated data collected under different environmental conditions.

\section{\methodnamenormal}\label{sec:fundamentals}

In this paper, we consider a model $f_\theta$ parametrized by $\theta$.
We would like our model to be robust or invariant to a set of transformations $\mathcal{T}$.
Formally, our goal is to find the model parameters $\theta$ that minimize the \emph{semantic adversarial risk}
\begin{equation}
\E_{(x,y) \sim \mathcal{D}} \left[ \maximize_{t \in \mathcal{T}} L(f_\theta(t(x)), y) \right],
\label{eq:risk}
\end{equation}
where $\mathcal{D} \subset \mathcal{X}\times\mathcal{Y}$ is a data distribution over pairs of examples $x$ and corresponding labels $y$, and $L$ is a suitable loss function (such as the $0-1$ loss in the context of classification tasks).
The set of semantic transformations $\mathcal{T}$ contains functions of the form $t : \mathcal{X} \rightarrow \mathcal{X}$.
Each element $t \in \mathcal{T}$ is irreducible and, crucially, for the optimal classifier $f_\theta : \mathcal{X} \rightarrow \mathcal{Y}$, we would like that $f_\theta(t(x)) = f_\theta(x)$ for all $t \in \mathcal{T}$.
For example, an \mnist classifier should not be affected by changes in the digit color.
In the following, we define a set of transformations $\mathcal{T}$ via a decoder that leverages a disentangled latent representation and explain how to evaluate the resulting risk in Equation~\eqref{eq:risk}.

\paragraph{Invariant latent factors.}

Disentanglement is perceived as a desirable property of representations.
Often, one hopes to obtain a representation of the observed data $x \in \mathcal{X}$ in terms of separate and conditionally independent factors $z \in \mathcal{Z}$ given $x$ under a certain class of input transformations \citep{higgins2018disentangled}. 
In our particular setting, we will assume a task-specific disentangled representation. Formally, we assume that we have an \emph{ideal} generator (or decoder), $\decode : \mathcal{Z} \rightarrow \mathcal{X}$, where the latent space $\mathcal{Z}$ is a product space of the form $\mathcal{Z} = \mathcal{Z}_{\parallel} \times \mathcal{Z}_{\perp}$. For a given classification task that predicts the label $y$, only the coordinates corresponding to $\mathcal{Z}_{\parallel}$ are relevant, while $\mathcal{Z}_{\perp}$ is irrelevant.
We formalize the above notions using conditional independence: given an example $x = \decode(z_{\parallel}, z_{\perp})$ with $z_{\perp} \in \mathcal{Z}_{\perp}$, $z_{\parallel} \in \mathcal{Z}_{\parallel}$ and corresponding label $y \in \mathcal{Y}$, we have
\begin{align}
    \mathbb{P}(y| z_{\parallel}, z_{\perp}) = \mathbb{P}(y| z_{\parallel}).\label{eq:invariance}
\end{align}
Hence, the ideal invariant classifier $f^\star$ that outputs a probability distribution over $\mathcal{Y}$ should be consistent with the invariance assumption
\begin{align}
f^\star(\decode(z_{\parallel}, z_{\perp})) = f^\star(\decode(z_{\parallel}, \tilde{z}_{\perp})) \label{eq:consistency}
\end{align}
for all $\tilde{z}_{\perp} \in \mathcal{Z}_{\perp}$, and should output the correct label:
\begin{align}
\argmax_{y' \in \mathcal{Y}} f^\star(\decode(z_{\parallel}, z_{\perp})) = y. \label{eq:correctness}
\end{align}

\noindent Finally, referring back to Equation~\eqref{eq:risk},
% and assuming that an \emph{ideal} disentangled representation $z = [z_{\parallel}, z_{\perp}]$ can be obtained from $x$ ($\exists z_{\parallel}, z_{\perp}$ s.t. $x = \decode(z_{\parallel}, z_{\perp})$),
we define the set of transforms $\mathcal{T}$ that induce semantically irrelevant perturbations as:
\begin{align}
    \mathcal{T} = \{ t ~|~ & t(x) =  \decode(z_{\parallel}, \tilde{z}_{\perp}) \textrm{ with  } \tilde{z}_{\perp} \in \mathcal{Z}_{\perp} \nonumber \\
    & \textrm{s.t.~} \exists z_{\perp} x = \decode(z_{\parallel}, z_{\perp}) \}. \label{eq:transforms}
\end{align}

\paragraph{Adversarial training.}

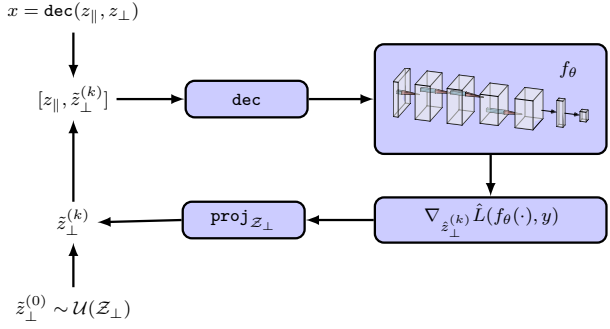
\begin{figure}
\centering
\resizebox{\linewidth}{!}{
\begin{footnotesize}
\begin{tikzpicture}[scale=1.]
    \node (input) [none] {$\tilde{z}^{(0)}_\perp \sim \mathcal{U}(\mathcal{Z}_\perp)$};
    \path (input.north)+(0, 1) node (zk) [none] {$\tilde{z}^{(k)}_\perp$};
    \path (zk.north)+(0, 1.58) node (latents) [none] {$[z_\parallel, \tilde{z}^{(k)}_\perp]$};
    \path (latents.north)+(0, 1) node (example) [none] {$x = \decode(z_\parallel, z_\perp)$};
    \path (latents.east)+(2, 0) node (dec) [op] {\decode};
    \path (dec.east)+(1, 0) node[anchor=west] (image) [op, text width=12em, minimum height=6em] {\includegraphics[width=3cm]{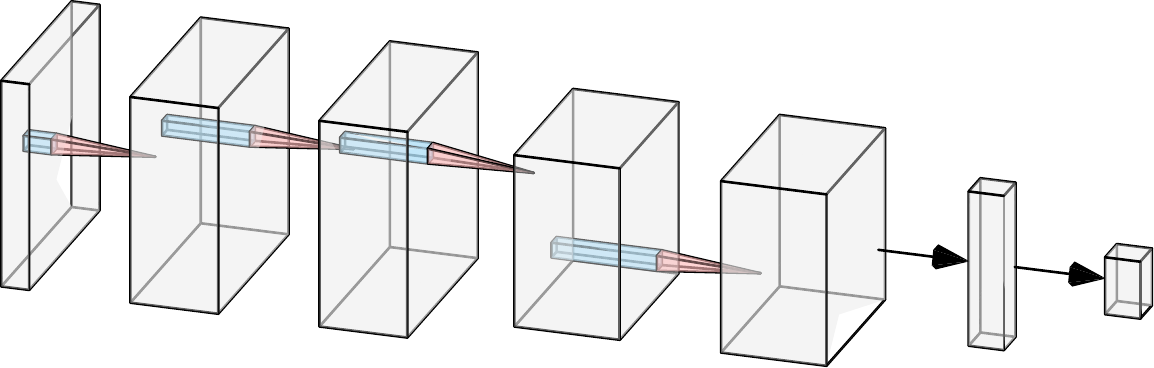}};
    \path (image.north)+(1.2, -.4) node (desc) [none] {$f_\theta$};
    \path (image.south)+(0, -1) node (grad) [op, text width=12em] {$\grad_{\hat{z}_\perp^{(k)}} \hat{L}(f_\theta(\cdot), y)$};
    \path (grad.west)+(-2, 0) node (proj) [op] {$\proj_{\mathcal{Z}_\perp}$};

    \path [draw, ->] (input.north) -- node [above] {} (zk.south);
    \path [draw, ->] (zk.north) -- node [above] {} (latents.south);
    \path [draw, ->] (example.south) -- node [above] {} (latents.north);
    \path [draw, ->] (latents.east) -- node [above] {} (dec.west);
    \path [draw, ->] (dec.east) -- node [above] {} (image.west);
    \path [draw, ->] (image.south) -- node [above] {} (grad.north);
    \path [draw, ->] (grad.west) -- node [above] {} (proj.east);
    \path [draw, ->] (proj.west) -- node [above] {} (zk.east);
\end{tikzpicture}
\end{footnotesize}
}
    
\caption{Illustration of the maximization process in Equation~\eqref{eq:max_rec}.}
    \label{fig:attack_latents}
\end{figure}

Given a model $f_\theta$ with enough capacity, minimizing the \emph{semantic adversarial risk} in Equation~\eqref{eq:risk} results in parameters $\theta^\star$
\begin{align}
    \theta^\star = \argmin_\theta \!\!\!\!\! \E_{\substack{(x,y) \sim \mathcal{D} \\ x = \decode(z_{\parallel}, z_{\perp})}} \left[ \maximize_{\tilde{z}_{\perp} \in \mathcal{Z}_\perp} L(f_\theta(\decode(z_{\parallel}, \tilde{z}_{\perp})), y) \right] \label{eq:opt_risk}
\end{align}
that satisfy Equations~\eqref{eq:consistency} and~\eqref{eq:correctness}.
In other words, there exists no transformation $t \in \mathcal{T}$ that, when applied to $x$, would result in a misclassification of the optimal classifier $f^\star = f_{\theta^\star}$.
Solving the saddle point problem in Equation~\eqref{eq:opt_risk} requires solving the corresponding inner-maximization problem
\begin{align}
    \tilde{z}_{\perp}^\star = \argmax_{\tilde{z}_{\perp} \in \mathcal{Z}_\perp} L(f_\theta(\decode(z_{\parallel}, \tilde{z}_{\perp})), y). \label{eq:zstar}
\end{align}
As enumerating all possible latents $\tilde{z}_{\perp} \in \mathcal{Z}_\perp$ is often intractable, we resort to a technique popularized by~\citet{madry_towards_2017} in the context of adversarial training,
which consists of using projected gradient ascent on a differentiable surrogate loss.
For a classification task, the $0-1$ loss is replaced with the cross-entropy loss:
\begin{align}
    \hat{L}(f_\theta(x), y) = - \log(\left [ f_\theta(x) \right]_y)    
\end{align}
where $[a]_i$ returns the i-th coordinate of $a$.
Gradient ascent steps are then interleaved with projection steps for a given number of iterations $K$.
Formally, we find an estimate $\tilde{z}^{(K)}_{\perp}$ of $\tilde{z}_{\perp}^\star$ using the following recursion:
\begin{align}
\tilde{z}^{(k+1)}_{\perp} = \proj_{\mathcal{Z}_\perp} \left( \tilde{z}_\perp^{(k)} + \alpha  \grad_{\tilde{z}_\perp^{(k)}} \hat{L}(f_\theta(\decode(z_{\parallel}, \tilde{z}^{(k)}_\perp)), y) \right) \label{eq:max_rec}
\end{align}
where $\tilde{z}_\perp^{(0)}$ is chosen at random within $\mathcal{Z}_\perp$, $\alpha$ is a constant step-size and $\proj_\mathcal{A}(a)$ is a projection operator that project $a$ onto $\mathcal{A}$.
Figure~\ref{fig:attack_latents} illustrates the process.

% \begin{figure}
%     \centering
%     \includegraphics[width=0.47\textwidth]{architectures.pdf}
%     \caption{Illustration of the maximization process in Equation~\eqref{eq:max_rec}.}
%     \label{fig:attack_latents}
% \end{figure}
% Attempting TikZ version.

Ultimately, \methodname (shortened as \methodabbrv) tries to find parameters that minimize the worst-case loss that could arise from altering the input examples through plausible transformations.
It guarantees that transformations of the input are meaningful by using a disentangled latent representation that encodes independent controllable factors, where some of these factors are known to be independent from the label.
Finding such a disentangled representation is rarely possible, as it is not always known which variations of the input should or should not affect the label.
In some cases, however, it is possible to train generative models such that we expect some subset of the latents to not affect the label.
Section~\ref{sec:method} implements \methodabbrv using a \stylegan model.

\begin{figure}[tb]
    \centering
    \includegraphics[width=0.47\textwidth]{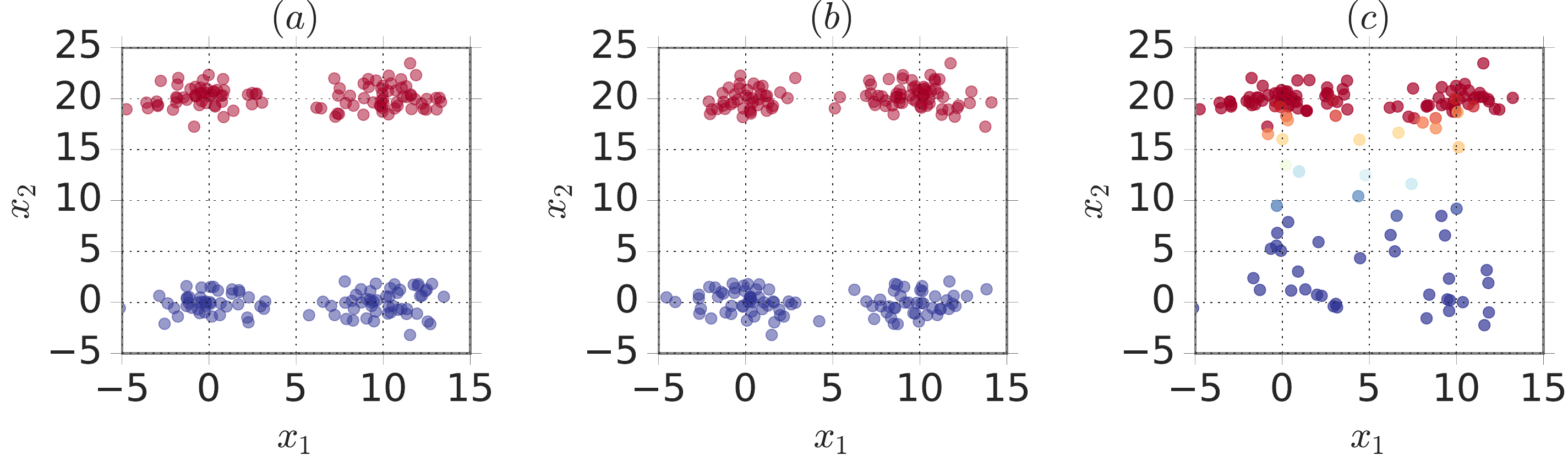}
    \caption{Comparison of \mixup and \methodabbrv on a toy example. In this example, we are given 200 datapoints. Each data point $(x_1, x_2)$ is sampled according to $x_1 \sim \mathcal{N}(z_{\perp}, \sqrt{3})$ where $z_{\perp} \in \mathcal{Z}_{\perp} = \lbrace 0., 10.\rbrace$ and $x_2 \sim \mathcal{N}(z_{\parallel}, 1)$ where $z_{\parallel} \in \mathcal{Z}_{\parallel} = \lbrace 0., 20.\rbrace$. The colors represent the label. Note that the latent variable $z_{\parallel} = 20y$ is dependent on the label while $z_{\perp}$ is independent of the label. Panel (a) shows the original set of 200 datapoints; panel (b) shows the effect of sampling additional  data using \methodabbrv; and panel (c) shows the effect of \mixup. Of course, we should point out that our method, \methodabbrv, is aware of the underlying latent representation, while \mixup is not.}
    \label{fig:data_augmentation}
\end{figure}

\paragraph{Data with low density regions.}
The motivation behind \methodabbrv stems from the \emph{manifold hypothesis}~\cite{fefferman2016testing}.
It states that high dimensional data present in the real-world, such as images, often lies on a low-dimensional manifold.
As a consequence, there exists large regions in the input space that are outside the support of the data distribution.
Hence, for maximal efficiency, \emph{data augmentation} and \emph{adversarial training} should be done carefully to make sure that the augmented data is still within the support of the original data distribution.
Data augmentation techniques presented in Figure~\ref{fig:augmentation_example} clearly violate this condition, and despite their success, we cannot expect that they perform well across all datasets (in fact, \mixup performs poorly on \coloredmnist).
Similarly, adversarial training targeting \lp-norm bounded perturbations tend to trade-off accuracy for robustness~\cite{ilyas2019adversarial}.
Figure~\ref{fig:data_augmentation} compares \mixup and \methodabbrv on a toy example.
In this example, we artificially construct a dataset with two classes and an underlying disentangled latent representation.
We observe that by exploiting the knowledge of the disentangled latent representation, \methodabbrv is capable of generating additional datapoints that are consistent with the original dataset, while \mixup generates additional datapoints that are unlikely.

\paragraph{Relationship to \mixup.}
\mixup augments data with respect to the input space.
Given two pairs of inputs $(x_A, y_A)$, $(x_B, y_B)$ and a linear interpolation factor sampled from a $\beta$-distribution $\lambda \sim \beta(\alpha, \alpha)$, \mixup generate a new input pair as follows:
\begin{align}
   \tilde{x} &= \lambda x_A + (1- \lambda) x_B \nonumber\\
   \tilde{y} &= \lambda y_A + (1 - \lambda) y_B.
\end{align}
Our methodology combines inputs $(x_A, y_A)$ and $(x_B, y_B)$ in the latent space.
If $x_A = \decode({z_A}_\parallel, {z_A}_\perp)$ and $x_B = \decode({z_B}_\parallel, {z_B}_\perp)$, we obtain
\begin{align}
  \tilde{x} &= \decode({z_A}_\parallel, {z_B}_\perp) \nonumber \\
  \tilde{y} &= y_A.
\end{align}
Crucially, this combination only affects the latent sub-space that is independent from the label, thus the label remains unchanged.
We also note that, unlike~\cite{verma2018manifold}, no interpolation occurs in the latent space (i.e.,  $\lambda {z_A}_\perp + (1-\lambda) {z_B}_\perp$) as this could result in points that are outside $\mathcal{Z}_\perp$ when $\mathcal{Z}_\perp$ is not convex.

\paragraph{Relationship to Invariant Risk Minimization.}

\citet{arjovsky2019invariant} consider the case where we have multiple datasets $D_e = \{x_i, y_i\}_{i=1}^n$ drawn from different training environments $e \in \mathcal{E}$.
As explained in~\cite{arjovsky2019invariant}, the motivation behind IRM is to minimize the worst-case risk
\begin{align}
    \maximize_{e \in \mathcal{E}} \E_{(x,y) \in D_e} \left[ L(f_\theta(x), y) \right]. \label{eq:irm}
\end{align}
In this paper, the environments are defined by the different instances of $z_{\perp} \in \mathcal{Z}_{\perp}$.
Given a dataset $\{ \decode({z_i}_\parallel, {z_i}_\perp), y_i \}_{i=1}^n$, we can rewrite the \emph{semantic adversarial risk} shown in Equation~\eqref{eq:risk} as Equation~\eqref{eq:irm} by setting the environment set $\mathcal{E}$ to
\begin{align}
    \mathcal{E} = \{ \{ \decode({z_i}_\parallel, z_\perp), y_i \}_{i = 1}^n | z_\perp \in \mathcal{Z}_{\perp} \}. \label{eq:irm_dataset}
\end{align}
This effectively create an ensemble of datasets for all possible combinations of $z_{\perp} \in \mathcal{Z}_{\perp}$ for all examples.

The crucial difference between IRM and \methodabbrv is in the formulation of the risk.
While IRM computes the risk by enumerating over a countable set of environments and picking the worst-case, \methodabbrv attempts to compute the worst-case risk by finding the combination of variations that maximize the risk over all examples.

\section{Implementation using \stylegan}\label{sec:method}

So far, we have assumed the presence of a generator (or decoder) that is capable of using a perfectly disentangled latent representation: we have assumed that this representation is partitioned into two subsets, one of which is known to be independent from the target label.
In practice, the methodology is often reversed: generative models are trained in the hope of obtaining some level of disentanglement.
If a partition of the trained latent space does not influence the label, we can use the corresponding trained generator within \methodabbrv.
This section explains why \stylegan is a good candidate and details how to implement \methodabbrv using \stylegan.
In particular, as we rely on \stylegan's \emph{mixing} property to enforce a partitioning of the latents, only three elements are needed:
\textit{(i)} a transformation set $\mathcal{Z}_\perp$ from which label-independent variants $\tilde{z}_\perp$ can be chosen;
\textit{(ii)} a dataset $\mathcal{D} = \{{z_i}_\parallel, y_i\}_{i = 1}^n$ of latents and labels; and
\textit{(iii)} an efficient method to find a worst-case variation $\tilde{z}_\perp \in \mathcal{Z}_\perp$.

\begin{figure*}[tb]
\centering
\begin{subfigure}{0.7\textwidth}
\centering
\includegraphics[height=1.3cm]{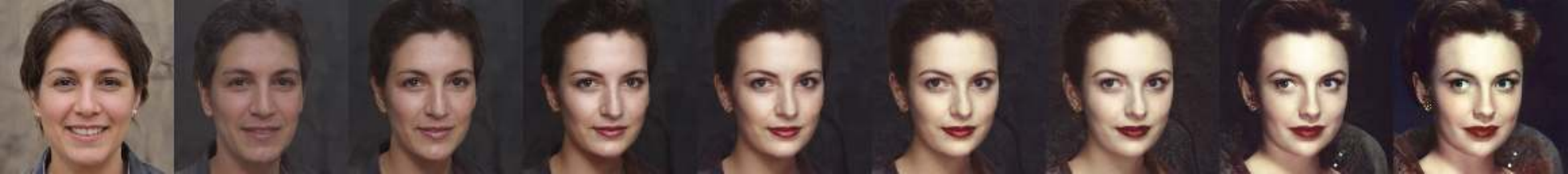}
\caption{\label{fig:encoder_example_process}}
\end{subfigure}
\hfill
\begin{subfigure}{0.25\textwidth}
\centering
\includegraphics[height=1.3cm]{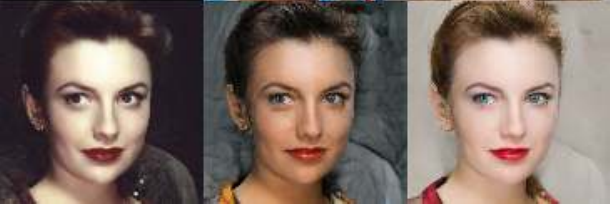}
\caption{\label{fig:encoder_example_mixed}}
\end{subfigure}
\caption{Panel \subref{fig:encoder_example_process} shows how the latents are progressively able to match a target image (on the far right). Panel \subref{fig:encoder_example_mixed} shows two different variations of the obtained image.
% The top row uses a \celeba target image, while the bottom row uses our \coloredmnist dataset.
\label{fig:encoder_example}}
\end{figure*}

\paragraph{\stylegan.}

\stylegan is a generator architecture for generative adversarial networks proposed by~\citet{karras2019style}.
It borrows interesting properties from the style transfer literature~\cite{huang2017arbitrary}. 
In this work, we rely on the style mixing property.
Formally, the \stylegan architecture is composed of two stages.
The first stage takes a latent variable $\mathtt{z}\sim\mathcal{N}(\vzero,\vone)$ that is not necessarily disentangled and projects it into a disentangled latent space $z = \map(\mathtt{z})$.
The second stage synthesizes an image $x$ from the disentangled latents $z$ using a decoder $x = \decode(z)$.
Overall, the process of generating an image $x$ using a \stylegan network is defined as
\begin{align}
x = \decode \circ \map (\mathtt{z}) ~\mbox{ where } \mathtt{z}\sim\mathcal{N}(\vzero,\vone).   
\end{align}
The intermediate latent variables $z$ provide some level of disentanglement that affects image generation at different spatial resolutions which allows us to control the synthesis of an image.
Particularly, we can apply the ``style'' of an image to another by mixing the disentangled latents of these images together.
% \stylegan allows us to have control over different resolutions of such styles.
In the context of face generation, the styles corresponding to coarse spatial resolutions affect high-level aspects such as pose, and styles of fine resolutions affect mainly the color scheme.
In the rest of this manuscript, we focus on variations of the finer style.\footnote{Other variations are possible as long as we have the certainty that they do not affect the label of interest.}
Concretely, our experiments in Section~\ref{sec:results} assume that the fine attributes $z_{\perp}$ are label-independent, while the coarse attributes $z_{\parallel}$ may be label-dependent.
Consequently, the finer style $ {z_B}_\perp$ of an image $x_\textrm{B}$ can be applied to another image $x_\textrm{A} = \decode({z_A}_\parallel, {z_A}_\perp)$ via $\decode(z_{\textrm{A}_{\parallel}}, z_{\textrm{B}_{\perp}})$.
Figure~\ref{fig:encoder_example_mixed} shows a nominal image and two variations of that image obtained by mixing the finer style of two other images.

\paragraph{Definition of the transformation set.}

For completeness, we now define the set of transforms $\mathcal{T}$ in Equation~\eqref{eq:transforms} by defining $\mathcal{Z}_\perp$.
While the formulation of \stylegan allows $\mathtt{z}$ to be sampled within an infinite support, our formulation requires $\mathcal{Z}_\perp$ to be bounded.
Additionally, as explained by~\citet{nalisnick2019detecting}, due to concentration of measure, a generative model usually draws samples from its typical set \cite{cover2012elements} (a subset of the model's full support) rather than regions of high probability density.\footnote{For $d$-dimensional isotropic Gaussian with standard deviation $\sigma$, the typical set resides at a distance of $\sigma\sqrt{d}$ from the mode~\cite{vershynin2018high}.}
As such, if $\mathtt{z} \in \R^d$, we wish to define $\mathcal{Z}_\perp$ as follows:
\begin{align}
    \!\!\!\! \mathcal{Z}_\perp = \left\{ \mathtt{map}(\mathtt{z})_\perp \bigg|\sqrt{d} - \delta d^{\frac{1}{4}} \leq \|\mathtt{z}\|_2 \leq \sqrt{d} + \delta d^{\frac{1}{4}} \right\} \label{eq:typicalz}
\end{align}
where $\delta$ is a small tunable positive constant.
In practice, however, we do not want to backpropagate through the \map operation as it is inefficient.
Instead, a small collection of latents is sampled, passed through the \map operation, and $\mathcal{Z}_\perp$ is limited to a neighborhood of the points in this collection.
This collection is re-sampled for each example and in expectation covers the typical set well (more details are given in Algorithm~\ref{alg:pgd}).

\paragraph{Construction of a dataset of disentangled latents.}

Constructing a dataset of labelled latents $\mathcal{D} = \{{z_i}_\parallel, y_i\}_{i = 1}^n$ requires finding the latents $z_i$ that decode into each example $x_i$ of an original labelled dataset $\{x_i, y_i\}_{i = 1}^n$.
Hence, we need to find a mapping between the image space and the latent space.
This mapping, which can be computed offline, is used to construct the dataset $\mathcal{D}$, and is only required once for each new dataset.
Specifically, this mapping is denoted as $\encode:\mathcal{X} \mapsto \mathcal{Z}$ and finds $z_i$ such that $x_i \approx \decode(z_i)$.
Algorithm~\ref{alg:encoder} defines this mapping through an optimization process.% (a Tensorflow excerpt is available in Appendix~\ref{sec:code}).
Inspired by~\cite{abdal2019image2stylegan}, and rather than relying solely on the distance between pixel values to define the loss of that optimization, we use the perceptual loss \citep{johnson2016perceptual,dosovitskiy2016generating} -- which helps steer the optimization process.
The perceptual loss is defined on the intermediate activations of a trained VGG-16 network~\citep{simonyan2014very} (see line~\ref{line:7}). 
We also found that the \stylegan generator, \decode, is a surjective mapping between its disentangled latent space and the image space (i.e., multiple latents can decode into the same image).
Hence, since we heavily rely on the mixing property of \stylegan, and to the contrary of \cite{abdal2019image2stylegan}, we propose to add an additional component to the loss that steers the latents towards a subset of latents that can be mixed.
In particular, we add a perceptual loss between the synthesized image and a mixed version of the same image (see lines~\ref{line:6} and~\ref{line:8}). Figure~\ref{fig:encoder_example} shows the evolution of the optimization process as well as mixed variants of the resulting image.

\begin{algorithm}[t]
\def\NoNumber#1{{\def\alglinenumber##1{}\State #1}\addtocounter{ALG@line}{-1}}
\footnotesize
\caption{Encoder \encode}
\begin{algorithmic}[1]
\Require Target image $x$, trained \stylegan model $\decode \circ \map$, and trained VGG network \vgg. $\alpha_i$ and $\beta_i$ are hyperparameters all set to 1 and 1/5 respectively. $\gamma^{(k)}$ is a step-size schedule.
\Ensure Disentangled latents $\hat{z}$ such that $\decode(\hat{z}) \approx x$
\State $\hat{z} \gets \frac{1}{M} \sum_{i=1}^M \map(\mathtt{z}^{(i)})$ with $\mathtt{z}^{(i)} \sim \mathcal{N}(\vzero,\vone)$ \Comment{Average latents}
\For{$k \in \{1, \ldots, N\}$} \Comment{$N$ is the number of iterations}
  \State $\hat{x} = \decode(\hat{z})$
  \State $\hat{\mathcal{A}} = \vgg(\hat{x})$ \Comment{\parbox[t]{.6\linewidth}{$\hat{\mathcal{A}}$ is a list of activations (after the 2$^{\textrm{nd}}$ convolution of 1$^{\textrm{st}}$, 2$^{\textrm{nd}}$ and 3$^{\textrm{rd}}$ blocks)}}
  \State $\mathcal{A} = \vgg(x)$
  \State $\mathcal{A}_\textrm{mix} = \vgg(\decode(\hat{z}_\parallel, \map(\mathtt{z})_\perp)))$ with $\mathtt{z} \sim \mathcal{N}(\vzero,\vone)$ \label{line:6}
  \State $L_\textrm{reconstruct} = \alpha_0 \| \hat{x} - x \|_2^2 + \sum_{i = 1}^{|\mathcal{A}|} \alpha_i \| \hat{\mathcal{A}}_i - \mathcal{A}_i \|_2^2$\label{line:7}
  \NoNumber{\Comment{Reconstruction loss}}
  \State $L_\textrm{mix} = \sum_{i = 1}^{|\mathcal{A}|} \beta_i \| \mathcal{A}_{\textrm{mix},i} - \mathcal{A}_{i} \|_2^2$ \Comment{Mixing loss} \label{line:8}
  \State $\hat{z} \gets \hat{z} - \gamma^{(k)} \grad_{\hat{z}} \left( L_\textrm{reconstruct} + L_\textrm{mix} \right)$
\EndFor
\end{algorithmic}
\label{alg:encoder}
\end{algorithm}

\paragraph{Generating worst-case examples to train robust models.}

\begin{algorithm}[t]
\footnotesize
\caption{Solution to Equation~\eqref{eq:zstar}}
\begin{algorithmic}[1]
\Require A nominal input $x$ and label $y$, a model $f_\theta$, a \stylegan model $\decode \circ \map$ and an encoder \encode. $L$ is the $0-1$ loss and $\hat{L}$ is the cross-entropy loss.
\Ensure Possible misclassified example $\tilde{x}$
\State $\tilde{x} \gets x$
\State $[z_\parallel, z_\perp] = \encode(x)$ \Comment{See Algorithm~\ref{alg:encoder}}
\For{$r \in \{1, \ldots, N_\textrm{r}\}$} \Comment{Repeat $N_\textrm{r}$ times}
  \State $\tilde{z}^{(0)}_\perp \gets \map(\mathtt{z})_\perp$ with $\mathtt{z} \sim \mathcal{N}(\vzero, \vone)$ \Comment{Initial latents}
  \State $\tilde{x}^{(0)} = \decode(z_\parallel, \tilde{z}^{(0)}_\perp)$
  \For{$k \in \{1, \ldots, K\}$} \Comment{K is the number of optimization steps}
    \State $\tilde{z}^{(k)}_\perp \gets \proj \left( \tilde{z}^{(k-1)}_\perp + \alpha \grad_{\tilde{z}^{(k-1)}_\perp} \hat{L}(f_\theta(\tilde{x}^{(0)}), y) \right)$
    \State $\tilde{x}^{(k)} = \decode(z_\parallel, \tilde{z}^{(k)}_\perp)$
    \If{$L(f_\theta(\tilde{x}^{(k)}), y) > L(f_\theta(\tilde{x}, y)$}
      \State $\tilde{x} \gets \tilde{x}^{(k)}$
      \State \Return \Comment{\parbox[t]{.6\linewidth}{Since $L$ is the $0-1$ loss, the procedure can terminate early}}
    \EndIf
  \EndFor
\EndFor
\end{algorithmic}
\label{alg:pgd}
\end{algorithm}

As explained in Section~\ref{sec:fundamentals}, minimizing the \emph{semantic adversarial risk} requires solving an inner-maximization problem.
We rely on projected gradient ascent on the cross-entropy loss $\hat{L}$ to efficiently find perturbed latents $\tilde{z}_\perp \in \mathcal{Z}_\perp$ such that, when mixed with $z_\parallel$, make the classifier output a label other than the true label.
Algorithm~\ref{alg:pgd} illustrates the process.
This algorithm approximates the typical set in Equation~\eqref{eq:typicalz} by randomly sampling initial latents $\tilde{z}^{(0)}_\perp$ $N_\textrm{r}$ times and projecting intermediate solutions $\tilde{z}^{(k)}_\perp$ back onto a neighborhood of $\tilde{z}^{(0)}_\perp$.\footnote{The actual implementation used in the experimental section projects back onto a \linf-bounded neighborhood around $\tilde{z}^{(0)}_\perp$: $\{ z_\perp | \| \tilde{z}^{(0)}_\perp - z_\perp \|_{\infty} < \epsilon \}$ where $\epsilon$ is set to 0.03.}
It refines the initial latents using gradient ascent with the goal of finding latents $\tilde{z}^{(K)}_\perp$ that, when mixed with the original image latents $z_\parallel$, generate an image $\decode(z_\parallel, \tilde{z}^{(K)}_\perp)$ that is misclassified.
Figure~\ref{fig:example} shows the result of this optimization procedure where the original image (on the top-left) is classified as ``not smiling'' and the optimized image (on the bottom-left) is classified as ``smiling''.
Once perturbed latents $\tilde{z}_\perp = \tilde{z}^{(K)}_\perp$ are found, we can compute the cross-entropy loss on the image generated by $\decode({z_i}_\parallel, \tilde{z}_\perp)$.
Formally, for a classifier $f_\theta$ and a dataset $\mathcal{D} = \{{z_i}_\parallel, y_i\}_{i = 1}^n$, we want to solve
\begin{align}
& \textrm{argmin}_\theta \expect{{z_i}_\parallel, y_i \sim D}{L(f_\theta(\decode({z_i}_\parallel, \tilde{z}_\perp)), y_i)} \label{eq:a} \\
& \textrm{and~} \tilde{z}_\perp = \textrm{argmax}_{z_\perp \in \mathcal{Z}_\perp} L(f_\theta(\decode({z_i}_\parallel, z_\perp)), y_i). \nonumber
\end{align}

\paragraph{Random mixing with disentangled representations.}

While this section describes an instantiation of \methodabbrv using \stylegan, it is possible to formulate an equivalent random \emph{data augmentation} baseline.
For an input $x$, we generate a random variation as follows:
\begin{align}
    \tilde{x} = \decode(\encode(x)_\parallel, \map(\mathtt{z})_\perp) \textrm{ with } \mathtt{z} \sim \mathcal{N}(\vzero, \vone)
\end{align}

\section{Results}\label{sec:results}

In this section, we compare \methodabbrv to {\it (i)} nominal training which minimizes the \emph{empirical risk}, {\it (ii)} Adversarial Training (AT) which minimizes the adversarial risk over \linf-norm bounded perturbations of size $\epsilon$ in input space~\cite{madry_towards_2017}, and {\it (iii)} Random Mixing with Disentangled Representations (\emph{RandMix}) which minimizes the \emph{vicinal risk} by randomly sampling latents from $\mathcal{Z}_\perp$ (rather than systematically finding the worst-case variations).
We perform two experiments to assess the generalization abilities of \methodabbrv.
The first experiment is done on an artificially constructed dataset called \coloredmnist (it bares resemblance to the \coloredmnist experiments present in~\cite{arjovsky2019invariant}).
The second experiment uses \celeba.
Both experiment demonstrate that methods using semantic variations as expressed by a trained \stylegan model achieve higher accuracy.
It also demonstrates that, when the distribution of variations is skewed (i.e., some variations $z_\perp$ appear more often than others in the dataset used to train the \stylegan model), \methodabbrv obtains higher accuracy than \randomabbrv.
For both experiments, we train a truncated VGG network with 5 layers using 5 epochs on \coloredmnist and 20 epochs on \celeba.
We use the Adam~\cite{kingma_adam:_2014} optimizer with a learning rate of $10^{-3}$.
\methodabbrv is trained with $N_\textrm{r}$ set to 5.

\subsection{\coloredmnist}

\begin{figure}[tb]
    \centering
    \includegraphics[width=0.47\textwidth]{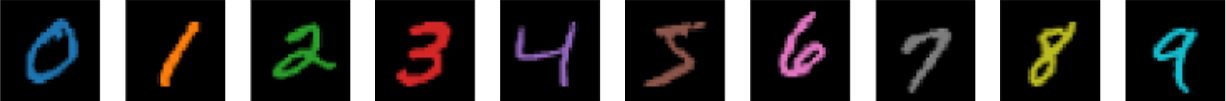}
    \caption{Mean colors given to each digit in the training set of our \coloredmnist case-study.}
    \label{fig:colored_mnist}
\end{figure}

\coloredmnist consists of a dataset of \mnist~\cite{lecun_mnist_2010} digits that are artificially colored to emphasize bias.
On the training set, we color each pair $(x, y)$ of the original \mnist dataset with a color drawn randomly from a normal distribution with mean $\mu_y$ and standard deviation $\sigma$ (means $\mu_y$ for $y \in \{0, \ldots, 9\}$ are shown in Figure~\ref{fig:colored_mnist}).
On the test set, we color digits uniformly at random.
In other words, the colors present in the training set spuriously correlate with the label.
We can use $\sigma$ to affect this correlation: by progressively increasing $\sigma$ the dataset becomes less biased.
For all techniques (including \mixup), we vary the level of bias and train models using 5 epochs.
The \stylegan model is trained on the training set only, once for each setting of $\sigma$.
The disentangled latents defining the finer style correspond to the final resolution of $32 \times 32$.\footnote{32 corresponds to the next power of two after 28 which is the size of an image in the original \mnist dataset.}

\begin{figure}[t]
\centering
\centering
\includegraphics[width=\columnwidth]{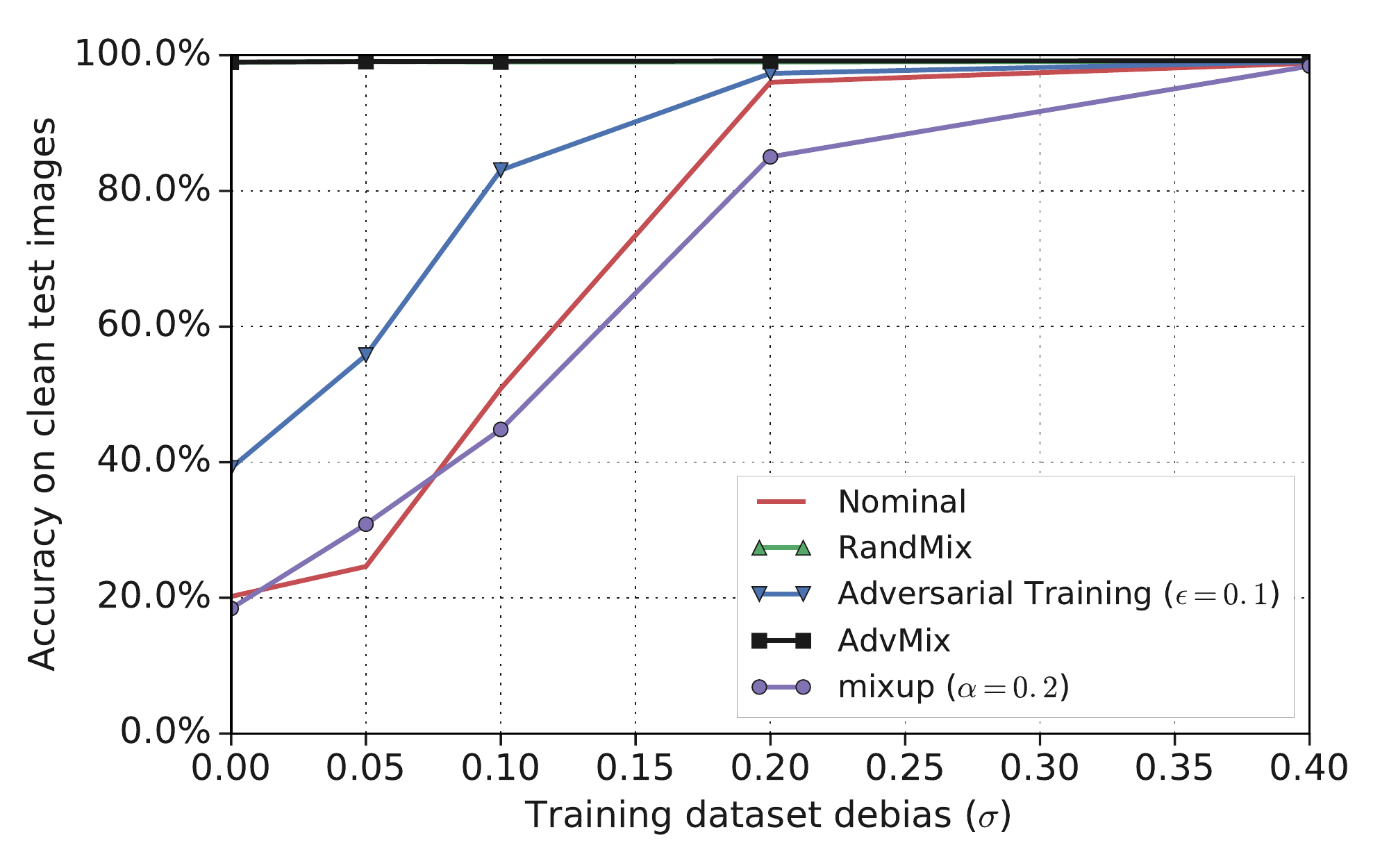}
\caption{
Accuracy of different training methods on images from our unbiased \coloredmnist test set.
The training set is progressively debiased by increasing the standard deviation of the colors present.
\label{fig:colored_mnist_biased_training}}
\end{figure}

Figure~\ref{fig:colored_mnist_biased_training} shows the results.
Across all settings, \randomabbrv and \methodabbrv outperform the other methods.
As expected, the gap between all methods decreases as the training set becomes less biased.
It is also worth noting that AT is useful (compared to nominal training and \mixup) as on this dataset \linf-norm bounded perturbations allow the exploration of slight variations in colors.
\randomabbrv and \methodabbrv are both expected to do well as all variations $z_\perp$ (that correspond to applications of different colors) are equally likely to be drawn from the \stylegan model (since they are uniformly distributed in the training set).

To further emphasize the difference between \randomabbrv and \methodabbrv, we purposefully bias the training of the \stylegan model.
We create two additional datasets (with $\sigma = 0$).
With the first dataset (named ``more biased''), the \stylegan model is trained on a large fraction of zeros (and few other digits), while on the second dataset (named ``less biased'', the \stylegan model is trained on a large fraction of zeros and ones.
As a result, rarely occurring variations (colors of digits from 1 to 9 for the first dataset and colors of digits from 2 to 9 for the second) are less likely to be randomly selected by \randomabbrv.
Table~\ref{table:biased1} shows the results.
We observe that \methodabbrv performs better.
However, we note that the gap is not large, as all color variations all contain red, green and blue components (which allows the network to implicitly learn about other color combinations).

\begin{table}[t]
\caption{
Effect of bias when training a \stylegan model on our \coloredmnist dataset.
\label{table:biased1}}
\vspace{-.5cm}
\begin{center}
\footnotesize
\begin{tabular}{l|ccc}
    \hline
    \cellcolor{Highlight} & \multicolumn{3}{c}{\cellcolor{Highlight} \bf Test accuracy on clean images} \Tstrut \\
    \cellcolor{Highlight} \textbf{Method} & \cellcolor{Highlight} Unbiased & \cellcolor{Highlight} Less biased & \cellcolor{Highlight} More biased  \Bstrut \\
    \hline
    \randomabbrv & 99.11\% & 98.87\% & 97.63\% \Tstrut \\
    \methodabbrv & \textbf{99.19\%} & \textbf{99.07\%} & \textbf{98.79\%} \Bstrut \\
    \hline
\end{tabular}
\end{center}
\end{table}

Finally, to create a stronger effect, we limit digits to the red, green and blue colors only (resulting in new datasets), and use a linear classifier (instead of a truncated VGG network).
Table~\ref{table:biased2} demonstrates that, when the \stylegan model is trained with a significant proportion of red digits, \methodabbrv does much better.
Indeed, \methodabbrv is able to systematically find the corner cases (i.e., green and blue variations) that are currently misclassified rather than relying on the random sampling of such cases.
We note that adversarial training can result in unstable learning, which can explain why \randomabbrv does slightly better when the \stylegan model is unbiased.

\begin{table}[t]
\caption{
Effect of bias when training a \stylegan model on our RGB \coloredmnist dataset (limited to red, blue or green colors). The classifier is a linear model (instead of a convolutional network).
\label{table:biased2}}
\begin{center}
\vspace{-.5cm}
\footnotesize
\begin{tabular}{l|ccc}
    \hline
    \cellcolor{Highlight} & \multicolumn{3}{c}{\cellcolor{Highlight} \bf Test accuracy on clean images} \Tstrut \\
    \cellcolor{Highlight} ~ & \cellcolor{Highlight} Unbiased & \cellcolor{Highlight} 99\% red & \cellcolor{Highlight} 99.9\% red \\
    \cellcolor{Highlight} \textbf{Method} & \cellcolor{Highlight}~ & \cellcolor{Highlight} Less biased & \cellcolor{Highlight} More biased \Bstrut \\
    \hline
    \randomabbrv & \textbf{88.55\%} & 83.18\% & 53.56\% \Tstrut \\
    \methodabbrv & 85.07\% & \textbf{85.02\%} & \textbf{85.00\%} \Bstrut \\
    \hline
\end{tabular}
\end{center}
\end{table}

\subsection{\celeba}

\celeba~\cite{liu2015deep} is a large-scale public dataset with forty different face attribute annotations including whether a person smiles or wears a hat.
We make no modifications to the dataset and use a pretrained \stylegan model~\footnote{https://github.com/NVlabs/stylegan}.
For all techniques, we train models using 20 epochs.
We evaluate all methods on their ability to classify the ``smiling'' attribute, as well as three other attributes.\footnote{Due to their sensitive connotation, we purposefully anonymized other attribute names and picked them from easier to harder classification tasks.}
In this experiment, the disentangled latents defining the finer style correspond to resolutions ranging from $128\times128$ to $1024\times1024$.\footnote{We always re-scale the resulting images to a $64\times64$ resolution.}

\begin{figure}[t]
\centering
\centering
\includegraphics[width=\columnwidth]{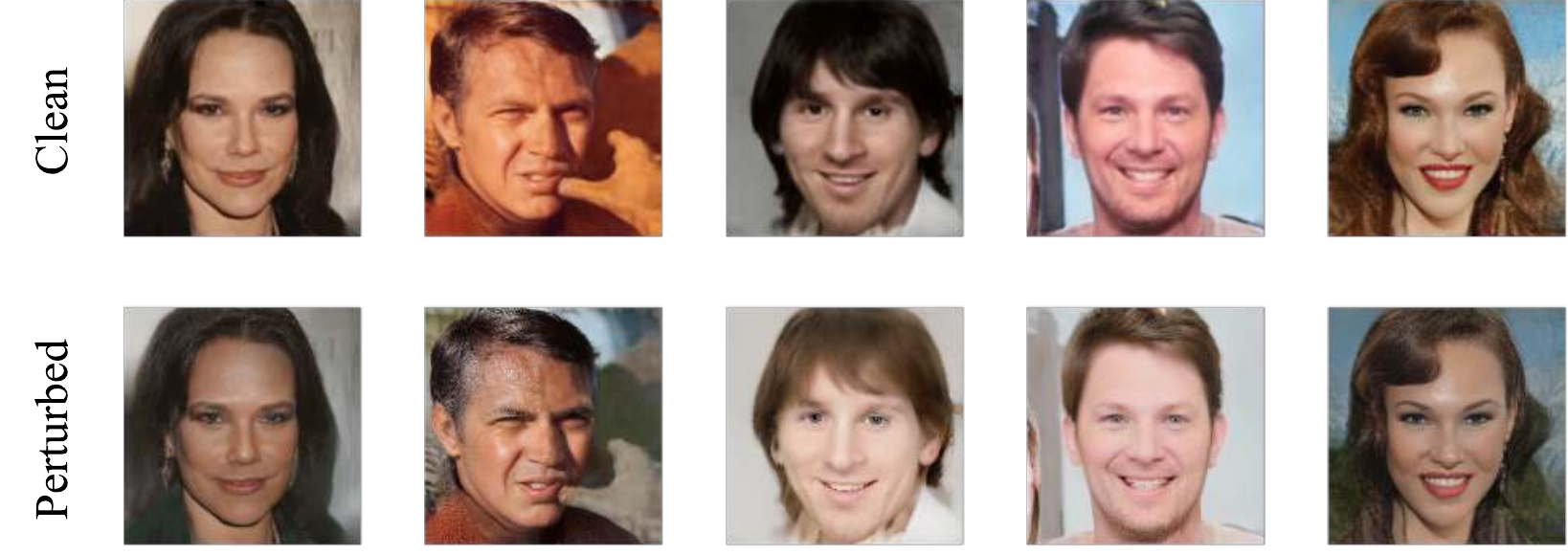}
\caption{
The top row shows examples of clean images from \celeba that are all classified correctly by the nominal model.
The bottom row shows semantically plausible variants of these images that are all misclassified.
\label{fig:misclassified}}
\end{figure}

In Table~\ref{table:results}, we observe that \methodabbrv is the only method that systematically achieves high accuracy.
This clearly demonstrates \methodabbrv can lead to a lower generalization error.
It is also interesting to see that \randomabbrv does not always improve on nominal training and that AT consistently trades off clean accuracy for \linf-robustness (as seen in \cite{ilyas2019adversarial}).
Finally, Figure~\ref{fig:misclassified} shows qualitative examples of images that are all correctly classified by the nominal model, but for which we can find plausible variants that are misclassified.
Appendix~\ref{sec:more_results} shows more results and includes other data augmentation schemes.

Overall, these results are confirming the observations made on the \coloredmnist dataset.
They seem to indicate that there is a slightly distributional shift between \celeba's train and test sets (at least when it comes to the finer image style).
By systematically probing variations that are difficult to classify, \methodabbrv is able to overcome this shift and reach better classification accuracy (to the contrary of \randomabbrv which can only stumble on difficult variants by chance).

\begin{table}[t]
\caption{
Test accuracy on different classification tasks of the \celeba dataset.
% attribute 1 = male
% attribute 2 = smiling
% attribute 3 = young
% attribute 4 = attractive
\label{table:results}}
\vspace{-.5cm}
\begin{center}
\footnotesize{
\begin{tabular}{l|cccc}
    \hline
    \cellcolor{Highlight} & \multicolumn{4}{c}{\cellcolor{Highlight} \bf Test accuracy on attribute} \Tstrut \\
    \cellcolor{Highlight} {\bf Method} & \cellcolor{Highlight} \#1 & \cellcolor{Highlight} \#2 (smiling) & \cellcolor{Highlight} \#3 & \cellcolor{Highlight} \#4 \Bstrut \\
    \hline
    Nominal & 96.49\% & 90.22\% & 83.52\% & 78.05\% \TBstrut \\
    \hline
    AT ($\epsilon = 4/255$) & 95.34\% & 91.11\% & 81.43\% & 76.61\% \Tstrut \\
    AT ($\epsilon = 8/255$) & 95.22\% & 89.29\% & 79.46\% & 74.39\% \Bstrut \\
    \hline
    \randomabbrv & 96.70\% & 90.36\% & 84.49\% & 76.41\% \Tstrut \\ 
    \methodabbrv & \textbf{97.56\%} & \textbf{92.29\%} & \textbf{85.65\%} & \textbf{79.47\%} \Bstrut \\
    \hline
\end{tabular}
}
\end{center}
\end{table}

\section{Conclusion}
We have demonstrated a novel approach to achieving robustness to input variations encountered in the real world by generating adversarial instances that compose disentangled representations.
We have shown how this framework can be realized by leveraging the \stylegan architecture -- resulting in 
models that are not only robust to systematic evaluation of insensitivity to variations but also exhibit better generalization, demonstrating that that accuracy is not necessarily at odds with robustness. 
Our formulation relies on good generative models that can learn a disentangled representation from which some directions are orthogonal to the label we are trying to predict.
Methods such as \methodabbrv are intended to be used to reduce the effect of bias and spurious correlations on classifiers.\footnote{It may be possible to use \methodabbrv to increase classification bias. However, other approaches are likely more effective to achieve this.}
We hope the promising results shown in this paper encourage the development of more effective disentangled representations that cover most factors of variations encountered in the real world. Finally, we hope this work leads to the exploration of this paradigm in the context of other Computer Vision applications and leads to the development of robust perception systems that can be safely used in the real world.

\clearpage
% Bibliography components
\setlength{\bibsep}{0.0pt}
{\small
\bibliographystyle{IEEEtranN}
\bibliography{bibliography}
}

\clearpage
\onecolumn
\appendix

\begin{center}
  {\Large \bf Achieving Robustness in the Wild via \\
Adversarial Mixing with Disentangled Representations \\
(Supplementary Material)}
\end{center}

\section{Additional examples}

\begin{figure*}[b]
\centering
\begin{subfigure}{.58\textwidth}
\centering
\includegraphics[width=\textwidth]{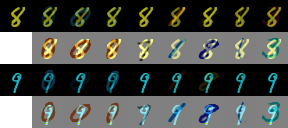}
\caption{\mixup}
\end{subfigure} \par\bigskip
\begin{subfigure}{.58\textwidth}
\centering
\includegraphics[width=\textwidth]{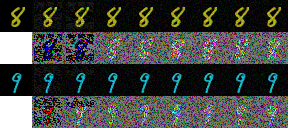}
\caption{Adversarial Training ($\epsilon=0.1$)}
\end{subfigure} \par\bigskip
\begin{subfigure}{.58\textwidth}
\centering
\includegraphics[width=\textwidth]{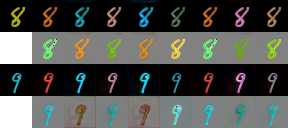}
\caption{\methodabbrv or \randomabbrv}
\end{subfigure}
\caption{
Example of perturbations obtained by different techniques on our \coloredmnist dataset.
The image on the far left is the original image.
On the same row are variations of that image.
Even rows show the rescaled difference between the original image and its variants.
\label{fig:mnist_extras}}
\end{figure*}

\begin{figure*}[b]
\centering
\begin{subfigure}{.6\textwidth}
\centering
\includegraphics[width=\textwidth]{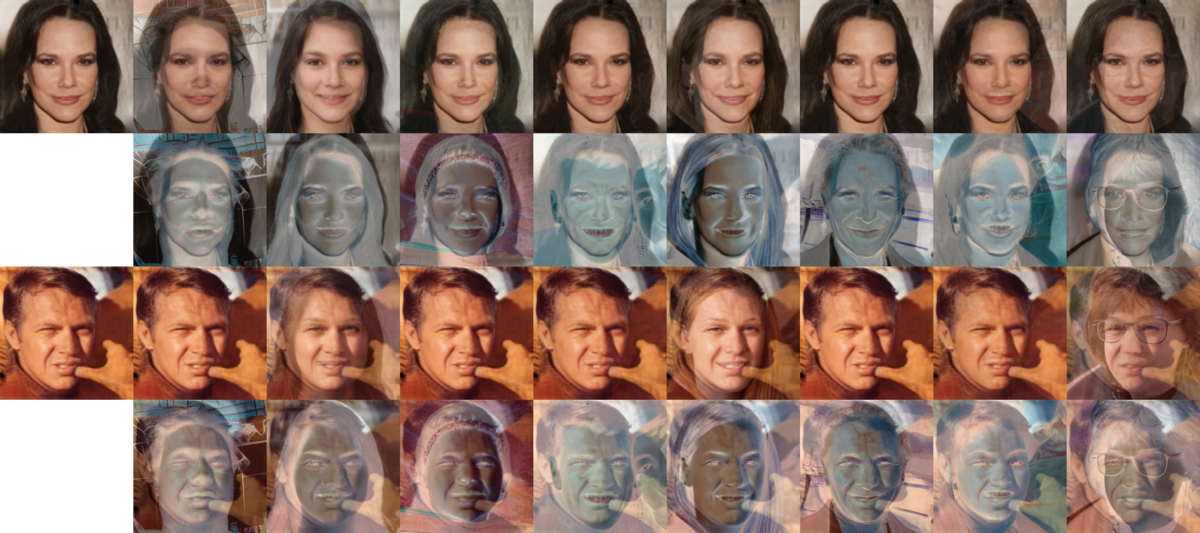}
\caption{\mixup}
\end{subfigure} \par\bigskip
\begin{subfigure}{.6\textwidth}
\centering
\includegraphics[width=\textwidth]{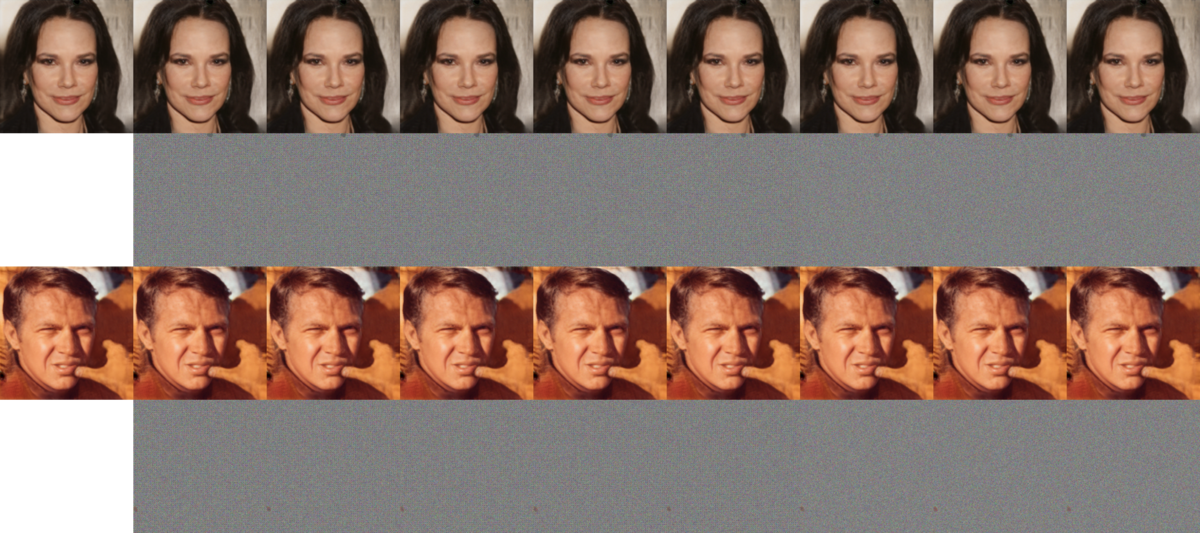}
\caption{Adversarial Training ($\epsilon=8/255$)}
\end{subfigure} \par\bigskip
\begin{subfigure}{.6\textwidth}
\centering
\includegraphics[width=\textwidth]{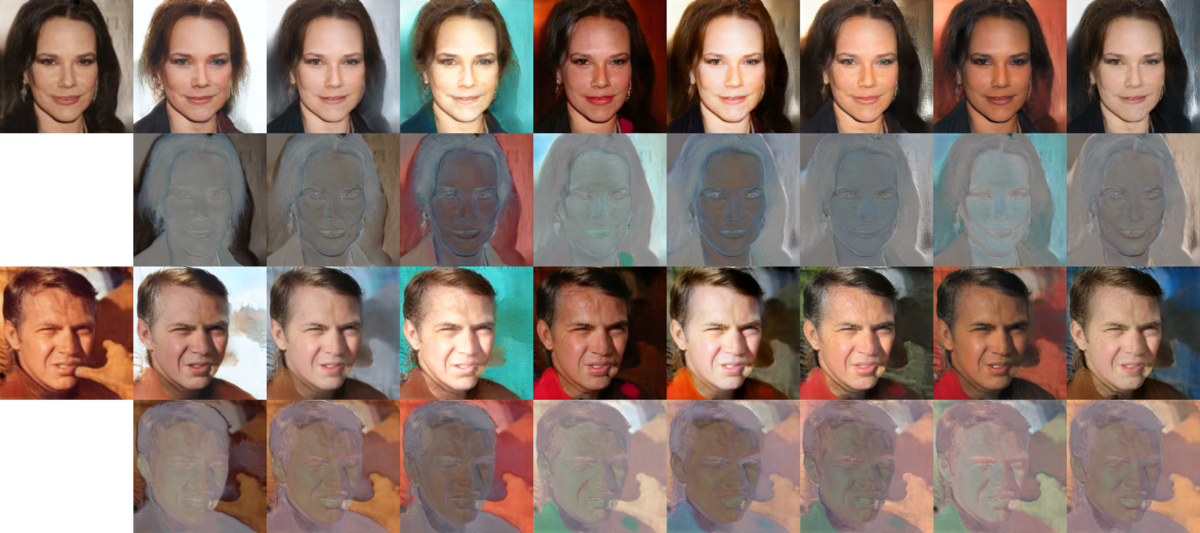}
\caption{\methodabbrv or \randomabbrv}
\end{subfigure}
\caption{
Example of perturbations obtained by different techniques on \celeba.
The image on the far left is the original image.
On the same row are variations of that image.
Even rows show the rescaled difference between the original image and its variants.
\label{fig:celeba_extras}}
\end{figure*}

Figure~\ref{fig:mnist_extras} shows additional examples of perturbations obtained on \coloredmnist by \textit{(a)} \mixup, \textit{(b)} adversarial attacks on \linf-bounded perturbations of size $\epsilon = 0.1$, and \textit{(c)} our method \methodabbrv.
Figure~\ref{fig:celeba_extras} shows examples on \celeba.
The underlying classifier is the nominally trained convolutional network.
We observe that the perturbations generated by \methodabbrv are semantically meaningful and result in plausible image variants -- to the contrary of the other two methods.

\begin{figure*}[h]
\centering
\includegraphics[width=.6\textwidth]{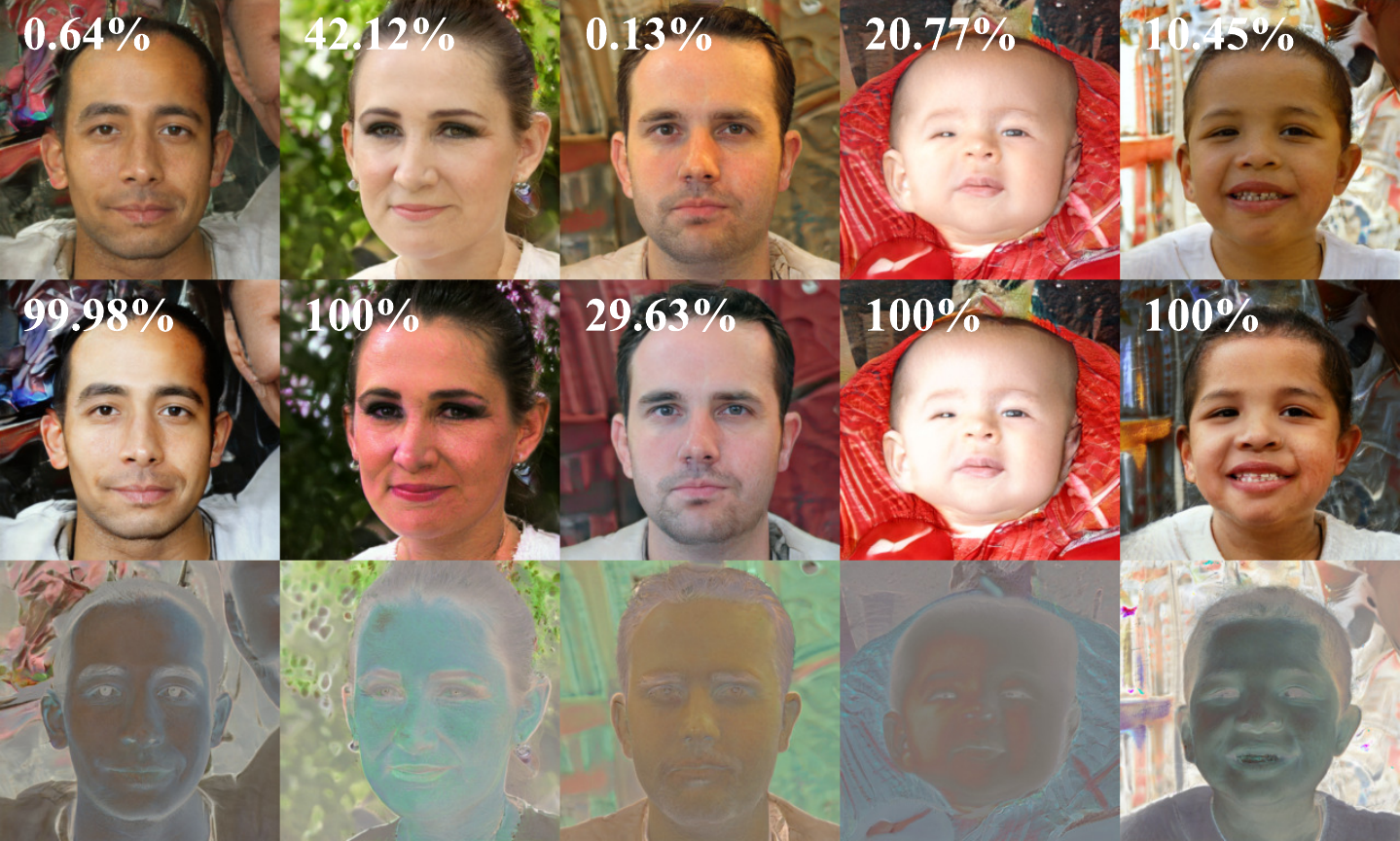}
\caption{
Example of perturbations obtained by \methodabbrv on randomly generated images.
The top row consists of images generated by a \stylegan model -- all these images are classified as ``not smiling'' by the nominal classifier (the numbers indicate the classifier output probability for ``smiling'').
The second row consists of adversarial perturbations obtained by \methodabbrv.
The last row shows the rescaled differences between the original images and their variant.
\label{fig:smile_attack}}
\end{figure*}

\clearpage
Figure~\ref{fig:smile_attack} shows image variants generated by \methodabbrv.
For four out of five images, \methodabbrv is able to change the  decision of a ``smile'' detector (nominally trained on \celeba).
We can qualitatively observe that brighter skin-tone and rosy cheeks tends to produce images that are more easily classified as ``smiling''.
Our interpretation is that pictures on the second row appear to be taken using flash photography (where it is more common for people to smile).
The second picture from the left (on the second row) also seem to be taken at night during an event.

\clearpage
\section{Additional results on \celeba}\label{sec:more_results}

For completeness, Table~\ref{table:mixup} shows the performance of \mixup, \cutout and \cutmix on the \celeba attributes used in Table~\ref{table:results}.
In addition to the evaluation of the unmodified clean test set, we also evaluate all methods by executing Algorithm~\ref{alg:pgd} with $N_\textrm{r}$ set to 10.
In other words, for each trained classifier, we try to find a misclassified variant for each example of the test set.
When a misclassified variant is found, we count the corresponding example as misclassified ``under perturbation''.

We observe that while data augmentation schemes such as \mixup, \cutout or \cutmix sometimes improve over nominal training, they do not provide consistent improvements.
With the exception of ``Attribute 3'' (where \cutmix is particularly efficient), \methodabbrv achieves the highest accuracy on the clean test set.
Additionally, as is expected, \methodabbrv systematically achieves the highest accuracy on perturbed images.

\begin{table}[h]
\caption{
Clean accuracy and accuracy under perturbations on different classification tasks of the \celeba dataset.
% attribute 1 = male
% attribute 2 = smiling
% attribute 3 = young
% attribute 4 = attractive
\label{table:mixup}}
\vspace{-.5cm}
\begin{center}
\footnotesize{
\begin{tabular}{l|cc}
    \hline
    \cellcolor{Highlight} & \multicolumn{2}{c}{\cellcolor{Highlight} \bf Accuracy on Attribute 1} \Tstrut \\
    \cellcolor{Highlight} \textbf{Method} & \cellcolor{Highlight} Clean & \cellcolor{Highlight} Under perturbation  \Bstrut \\
    \hline
    Nominal & 96.49\% & 40.35\% \TBstrut \\
    \hline
    \mixup $(\alpha = 0.2)$ & 97.22\% & 50.48\% \Tstrut \\
    \cutout & 96.92\% & 61.58\% \\
    \cutmix & 97.18\% & 32.86\% \Bstrut \\
    \hline
    AT \linf with $\epsilon = 4/255$ & 95.34\% & 48.91\% \Tstrut \\
    AT \linf with $\epsilon = 8/255$ & 95.22\% & 45.01\% \Bstrut \\
    \hline
    \randomabbrv & 96.70\% & 39.41\% \Tstrut \\
    \methodabbrv & \textbf{97.56\%} & \textbf{84.29\%} \Bstrut \\
    % Semantic AT ``strong'' & 97.29\% & 81.20\% \Bstrut \\
    \hline
    \multicolumn{3}{c}{\vspace{-3mm}} \\
    \hline
    \cellcolor{Highlight} & \multicolumn{2}{c}{\cellcolor{Highlight} \bf Accuracy on Attribute 2 (smiling)} \TBstrut \\
    \hline
    Nominal & 90.22\% & 18.60\% \TBstrut \\
    \hline
    \mixup $(\alpha = 0.2)$ & 90.95\% & 30.49\% \Tstrut \\
    \cutout & 90.44\% & 17.55\% \\
    \cutmix & 90.88\% & 15.46\% \Bstrut \\
    \hline
    AT \linf with $\epsilon = 4/255$ & 91.11\% & 60.93\% \Tstrut \\
    AT \linf with $\epsilon = 8/255$ & 89.29\% & 56.19\% \Bstrut \\
    \hline
    \randomabbrv & 90.36\% & 23.51\% \Tstrut \\
    \methodabbrv & \textbf{92.29\%} & \textbf{74.55\%} \Bstrut \\
    % Semantic AT ``strong'' & 91.42\% & 71.04\% \Bstrut \\
    \hline
    \multicolumn{3}{c}{\vspace{-3mm}} \\
    \hline
    \cellcolor{Highlight} & \multicolumn{2}{c}{\cellcolor{Highlight} \bf Accuracy on Attribute 3} \TBstrut \\
    \hline
    Nominal & 83.52\% & 3.31\% \TBstrut \\
    \hline
    \mixup $(\alpha = 0.2)$ & 85.16\% & 3.51\% \Tstrut \\
    \cutout & 84.94\% & 2.91\% \\
    \cutmix & \textbf{85.67\%} & 1.47\% \Bstrut \\
    \hline
    AT \linf with $\epsilon = 4/255$ & 81.43\% & 52.92\% \Tstrut \\
    AT \linf with $\epsilon = 8/255$ & 79.46\% & 62.71\% \Bstrut \\
    \hline
    \randomabbrv & 84.49\% & 3.19\% \Tstrut \\
    \methodabbrv & 85.65\% & \textbf{69.55\%} \Bstrut \\
    % Semantic AT ``strong'' & 85.41\% & 62.61\% \Bstrut \\
    \hline
    \multicolumn{3}{c}{\vspace{-3mm}} \\
    \hline
    \cellcolor{Highlight} & \multicolumn{2}{c}{\cellcolor{Highlight} \bf Accuracy on Attribute 4} \TBstrut \\
    \hline
    Nominal & 78.05\% & 0.23\% \TBstrut \\
    \hline
    \mixup $(\alpha = 0.2)$ & 76.80\% & 0.03\% \Tstrut \\
    \cutout & 76.59\% & 0.14\% \\
    \cutmix & 78.50\% & 0.19\% \Bstrut \\
    \hline
    AT \linf with $\epsilon = 4/255$ & 76.61\% & 9.74\% \Tstrut \\
    AT \linf with $\epsilon = 8/255$ & 74.39\% & 5.68\% \Bstrut \\
    \hline
    \randomabbrv & 76.41\% & 0.42\% \Tstrut \\
    \methodabbrv & \textbf{79.47\%} & \textbf{47.95\%} \Bstrut \\
    % Semantic AT ``strong'' & 78.03\% & 42.23\% \Bstrut \\
    \hline
\end{tabular}
}
\vspace{-.5cm}
\end{center}
\end{table}

\clearpage
\section{Code snippets}\label{sec:code}

This section shows how to implement Algorithms~\ref{alg:encoder} and~\ref{alg:pgd} in TensorFlow~2~\cite{abadi_tensorflow:_2016}. Below is Algorithm~\ref{alg:encoder}.

\begin{lstlisting}[language=Python]
import numpy as np
import tensorflow as tf

import imagenet  # Custom package which contains `VGG16`.
import stylegan  # Custom package which contains `Generator`.


# Hyper-parameters.
num_steps = 2000
def learning_rate_schedule(t):
    if t < 1500:
        return .05
    return .01
mixing_level = 8
    
# Optimizer.
learning_rate = tf.Variable(0., trainable=False, name='learning_rate')
optimizer = tf.keras.optimizers.Adam(learning_rate)

# Create the StyleGAN generator.
# `generator` has a the following properties:
# - latent_size: Number of latent coordinates (for the FFHQ model, this is equal to 512)
# - average_disentangled_latents: Average disentangled latents.
# It also has the following functions:
# - map(z) which runs the mapping operation that transforms latents into disentangled latents.
# - synthesize(z_disentangled) which generates an image from disentangled latents.
generator = stylegan.Generator()

# Create a VGG network.
# `vgg` has the following function:
# - get_activations(x) which computes the VGG activations at its 1st block 2nd convolution,
#   3rd block 2nd convolution, 4th block 2nd convolution.
vgg = imagenet.VGG16()

# Target image.
target_image = load_target_image()
activations_of_target_image = [target_image] + vgg(tf.image.resize(target_image, 225, 225))

# `disentangled_latents` will be optimized to match `target_image`.
# For the FFHQ model, `disentangled_latents` has shape [18, 512].
disentangled_latents = tf.Variable(average_disentangled_latents, name='disentangled_latents')


def feature_loss(a, b):
    s = float(np.prod(a.shape.as_list()))
    return tf.reduce_sum(tf.square(a - b)) / s


def compute_loss(z):
    """Computes the loss."""
    generated_image = generator.synthesize(z)
    activations_of_generated_image = (
        [generated_image] + vgg(tf.image.resize(generated_image, 225, 225)))
    
    # Reconstruction and perceptual loss.
    zipped_activations = zip(activations_of_generated_image, activations_of_target_image)
    loss = sum([feature_loss(g, t) for g, t in zipped_activations])
    
    # Generated a randomly mixed image.
    random_latents = tf.random.normal(shape=(generator.latent_size,))
    random_disentangled_latents = generator.map(random_latents)
    mixed_disentangled_latents = tf.concat([
        z[:mixing_level, :],
        random_disentangled_latents[mixing_level:, :]
    ], axis=0)
    mixed_image = generator.synthesize(mixed_disentangled_latents)
    
    # Perceptual loss on mixed image.
    activations_of_mixed_image = vgg(tf.image.resize(mixed_image, 225, 225))
    zipped_activations = zip(activations_of_mixed_image, activations_of_target_image[1:])
    loss += sum([.2 * feature_loss(g, t) for g, t in zipped_activations])
    
    return loss


# Run the optimization procedure.
for step in range(num_steps):
    with tf.GradientTape() as tape:
        tape.watch(disentangled_latents)
        loss = compute_loss(disentangled_latents)
    grads = tape.gradient(loss, [disentangled_latents])
    learning_rate.assign(learning_rate_schedule(step))
    optimizer.apply(grads, [disentangled_latents])
\end{lstlisting}

Below is Algorithm~\ref{alg:pgd}.

\begin{lstlisting}[language=Python]
import tensorflow as tf

import stylegan  # Custom package which contains `Generator`.


# Hyper-parameters.
mixing_level = 8
epsilon = .03
num_restarts = 5  # Or 10 for evaluation (the higher the better).
num_steps = 7  # Or 20 for evaluation.
step_size = .005

# Create the StyleGAN generator.
# `generator` has a the following properties:
# - latent_size: Number of latent coordinates (for the FFHQ model, this is equal to 512)
# - average_disentangled_latents: Average disentangled latents.
# It also has the following functions:
# - map(z) which runs the mapping operation that transforms latents into disentangled latents.
# - synthesize(z_disentangled) which generates an image from disentangled latents.
generator = stylegan.Generator()

# `classifier` is the current classification model we are training or evaluating.
# `classifier` has the following function:
# - get_logits(x) which returns the logits of x.
classifier = load_current_model()

# `z` are precomputed latents obtained through Algorithm 1.
# For the FFHQ model, `z` has shape [18, 512].
# `y` is the class of the original image that can be synthesized through z.
z = load_disentangled_latents()
z_parallel = z[:mixing_level, :]
y = get_class_of(z)


def project(x, x0, eps):
    return x0 + tf.clip_by_value(x - x0, -eps, eps)


# Initialization.
x = generator.synthesize(z)
logits = classifier.get_logits(x)
highest_loss = tf.nn.sparse_softmax_cross_entropy_with_logits(y, logits)
adversarial_x = x

for _ in range(num_restarts):
    # Pick a random set of disentangled latents.
    z = tf.random.normal(shape=(generator.latent_size,))
    z_perp = z_perp_original = generator.map(z)[mixing_level:, :]
    
    for _ in range(num_steps):
        with tf.GradientTape() as tape:
            tape.watch(z_perp)
            # Mix latents.
            mixed_z = tf.concat([z_parallel, z_perp], axis=0)
            # Generate and classify.
            x = generator.synthesize(mixed_z)
            logits = classifier.get_logits(x)
            # Cross-entropy loss.
            loss = tf.nn.sparse_softmax_cross_entropy_with_logits(y, logits)
        if loss > highest_loss:
            adversarial_x = x
            highest_loss = loss
            
        grad = tape.gradient(loss, [z_perp])[0]
        z_perp += step_size * tf.math.sign(grad)  # Gradient ascent.
        z_perp = project(z_perp, z_perp_original, epsilon)
        
    # Last step.
    mixed_z = tf.concat([z_parallel, z_perp], axis=0)
    x = generator.synthesize(mixed_z)
    logits = classifier.get_logits(x)
    loss = tf.nn.sparse_softmax_cross_entropy_with_logits(y, logits)
    if loss > highest_loss:
        adversarial_x = x
        highest_loss = loss
\end{lstlisting}

\end{document}